\pdfoutput=1

\documentclass[11pt]{article}

\usepackage[final]{EMNLP2023}

\usepackage{times}
\usepackage{latexsym}

\usepackage[T1]{fontenc}

\usepackage[utf8]{inputenc}

\usepackage{microtype}

\usepackage{inconsolata}

\usepackage{amsfonts,amssymb}

\usepackage{amsmath}
\usepackage{cleveref}
\usepackage{titlesec}
\crefname{section}{§}{§§}
\Crefname{section}{§}{§§}
\usepackage[utf8]{inputenc} 
\usepackage[T1]{fontenc}    
\usepackage{hyperref}       
\usepackage{url}            
\usepackage{booktabs}       
\usepackage{multirow}
\usepackage{tabularx}
\usepackage{bm} 
\usepackage{amsfonts}       
\usepackage{nicefrac}       
\usepackage{microtype}      
\usepackage{xcolor}         
\usepackage{subfigure}
\usepackage{makecell}

\usepackage[tikz]{bclogo}
\definecolor{msftBlue}{RGB}{0,164,239}
\definecolor{msftGreen}{RGB}{127,186,0}
\definecolor{msftYello}{RGB}{255,185,0}
\definecolor{msftBlack}{RGB}{0,0,0}

\definecolor{midnightgreen}{rgb}{0.0, 0.29, 0.33}
\definecolor{deepgreen}{HTML}{0aa344}

\usepackage{enumitem}

\newenvironment{itemize*}%
 {\leftmargini=20pt\begin{itemize}%
  \setlength{\itemsep}{3pt}%
  \setlength{\parskip}{0pt}%
  }%
 {\end{itemize}}
\newenvironment{enumerate*}%
 {\begin{enumerate}%
  \setlength{\itemsep}{0pt}%
  \setlength{\parskip}{0pt}}%
 {\end{enumerate}}

\newcommand{\red}{\textcolor{red}}
\newcommand{\green}{\textcolor{deepgreen}}
\usepackage{soul}
\usepackage{utfsym}
\usepackage{hyperref}
\usepackage{cleveref}
\usepackage{chemarrow}

\def \framework{Toolink}

\title{{\framework}: Linking Toolkit Creation and Using through Chain-of-Solving on Open-Source Model}

\author{
 Cheng~Qian$^{1}$, Chenyan~Xiong$^{2}$, Zhenghao~Liu$^{3}$, Zhiyuan~Liu$^{1}$\\
 $^1$Tsinghua University,\hspace{0.3em}$^2$Carnegie Mellon University,\hspace{0.3em}$^3$Northeastern University \\
\texttt{qianc20@mails.tsinghua.edu.cn}\\
}

\begin{document}
\maketitle

\begin{abstract}
Large Language Models (LLMs) have demonstrated remarkable progress in utilizing tools, but their closed-source nature and high inference costs pose limitations on their adaptability, necessitating a valid method that leverages smaller, open-sourced models. In this paper, we introduce Toolink, a comprehensive framework that performs task-solving by first creating a toolkit and then integrating the planning and calling of tools through a chain-of-solving (CoS) approach. We first validate the efficacy of Toolink in harnessing the model's creativity and CoS ability on ChatGPT. Subsequently, we curate CoS-GPT, a chain-of-solving dataset designed for tool-using, and finetune the LLaMA-7B model. It results in LLaMA-CoS, a powerful open-source model with advanced tool-planning and tool-calling capabilities. Evaluation of diverse tasks from BIG-bench demonstrates its CoS ability matches that of ChatGPT while its performance surpasses the chain-of-thought approach. Further studies highlight the generalization of LLaMA-CoS to unseen tasks and showcase its capability in using toolkits not explicitly tailored for the target task, affirming its robustness in real-world scenarios. All codes and data are released\footnote{\url{https://github.com/qiancheng0/Toolink}}.
\end{abstract}

\section{Introduction}

\begin{figure}[t]
    \centering
    \includegraphics[width=1.02\linewidth]{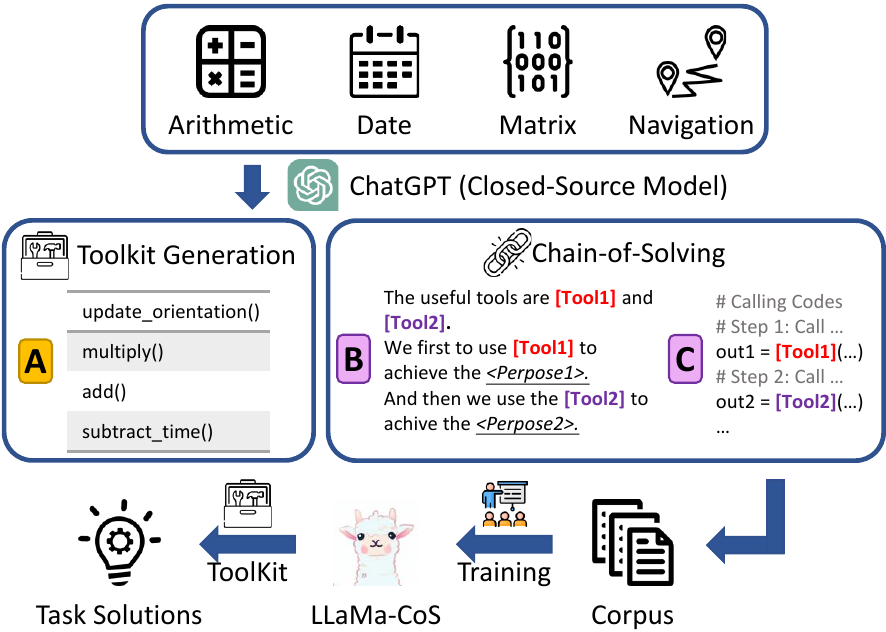}
    \caption{An illustration of Toolink, which decomposes tasks via toolkit creation and resolves queries through Chain-of-Solving (CoS). Toolink can be adapted to open-source LLaMA for enhanced tool usage.}
    \label{fig:brief_pipeline}
\end{figure}

Large Language Models (LLMs) such as Codex~\citep{chen2021evaluating}, ChatGPT~\citep{openai2022chatgpt}, and GPT4~\citep{openai2023gpt4} have made significant strides in code generation, in-context learning, and logical reasoning. However, they still struggle with precise calculations and accessing current information~\citep{patel2021nlp, trivedi2022interleaving, lu2022survey}. To address these issues, research has focused on equipping LLMs with tools such as calculators~\citep{cobbe2021training, parisi2022talm, schick2023toolformer}, search engines~\citep{carlini2021extracting, thoppilan2022lamda, schick2023toolformer}, scratch pads~\citep{nye2021show}, calendars~\citep{schick2023toolformer}, and image retrievers~\citep{sheynin2022knn} to enhance their capabilities, thus benefiting various tasks including question-answering, math calculations, and long-form generation. Recent studies have also explored how LLMs can devise plans, make decisions, and perform tool invocations~\citep{shen2023hugginggpt, lu2023chameleon, liang2023taskmatrix}. By combining them into a pipeline, these frameworks aim to construct more advanced NLP systems for improved task performance.

However, current tool-using pipelines heavily rely on closed-source models with inaccessible parameters. It poses challenges particularly as follows: i) \textbf{Limited adaptability}: The closed-source nature of major LLMs prevents them from customization, resulting in a lack of flexibility to adapt to tasks with specific requirements. ii) \textbf{Low efficiency and high inference cost}: Many existing LLMs can only be accessed \textit{online}, which imposes limitations on the inference rate and leads to high expense. iii) \textbf{Privacy and security concerns}: Each query must be submitted to these closed-source LLMs to obtain a tool-using solution, which raises concerns regarding potential privacy breaches and compromises data security.

To address these challenges, we propose Toolink, a comprehensive framework to boost the tool-using ability of open-source models with the help of closed-source models. As shown in \Cref{fig:brief_pipeline}, Toolink first decomposes the target task by creating a toolkit for problem-solving, and then leverages the open-source model to use tools to answer queries in a chain-of-solving (CoS) approach. Specifically, CoS disentangles the model's reasoning through two stages: \textit{CoS-Planning}, which selects useful tools from the created toolkit and plans their usages based on the specific query; and \textit{CoS-Calling}, which focuses on deriving the answer by performing tool invocations in code format according to the plan devised. To effectively train the open-source model in these abilities, we employ ChatGPT to curate CoS-GPT, a training dataset that aims to inspire the tool-using ability of open-source models through CoS. Specifically, we finetune LLaMA-7B~\citep{touvron2023llama} into LLaMA-CoS, which is equipped with strong tool-using capabilities by linking toolkit creation with the chain of problem-solving.

LLaMA-CoS can solve the queries \textit{offline} without uploading queries to closed-source models, ensuring data security and privacy. Experiments further illustrate that Toolink outperforms the chain-of-thought (CoT)~\citep{weichain} on diverse tasks from BIG-bench~\citep{srivastava2022imitation} and enables LLaMA-CoS to showcase comparable CoS ability to that of ChatGPT. In addition, LLaMA-CoS can generalize to unseen tasks by planning and calling tailored tools, and solve the target task with a toolkit not specifically tailored for it. These findings further affirm our framework's robustness in solving queries under real-world scenarios.

\section{Related Work}
\paragraph{Tool-based enhancement for LLMs.} Language models have been enhanced with external tools to improve their expertise. Previous work focused on equipping the LLMs with different tools including a calculator to improve calculation accuracy~\citep{cobbe2021training, parisi2022talm, schick2023toolformer}, search engine to inquire factual knowledge~\citep{carlini2021extracting, thoppilan2022lamda, schick2023toolformer}, Python interpreter to execute programs~\citep{chen2022program, gao2022pal}, and retriever to search textual information~\citep{khandelwalgeneralization, borgeaud2022improving}, etc.

More recent studies, such as HuggingGPT~\citep{shen2023hugginggpt}, Chameleon-LLM~\citep{lu2023chameleon}, VisualGPT~\citep{wu2023visual} and TaskMatrix.AI~\citep{liang2023taskmatrix}, focus on assembling plannings, execution, and reasoning about tools into a robust pipeline. In addition to tool-using, ART~\citep{paranjape2023art} builds toolkits based on retrieved tasks from the manually built library, while LATM~\citep{cai2023large} and CREATOR~\citep{qian2023creator} involve the LLMs' tool-making ability to offload their reasoning burden and raise task performance. In contrast to their prevalent use of closed-source LLMs to leverage tools, {\framework} offers unique advantages of tool use for smaller, open-source models.

\paragraph{Adaptation of open-source models.} One research direction focuses on effective tuning of open-source models, including the introduction of lightweight modules such as Adapter~\citep{houlsby2019parameter} and LoRA~\citep{hu2021lora}. These modules are adapted to various model types including LLaMA~\citep{touvron2023llama}, T5~\citep{raffel2020exploring}, and other Transformers-based architectures~\citep{pfeiffer2020adapterhub}, to save computational resources. For instance, GOAT~\citep{liu2023goat} applies LoRA to improve LLaMA's arithmetic calculation ability, while LLaMA-Adapter~\citep{zhang2023llama} adopts Adapter and zero-init attention to improve multi-modal task performance.

Other works have investigated how instruction tuning can make open-source models better understand and follow human requirements in both text format~\citep{longpre2023flan,ouyang2022training} and visual domains~\citep{liu2023visual}. More recent works also investigate the curation of instruction following data~\citep{stanford_alpaca, peng2023instruction} and construction of open-source tool-using agents~\citep{qin2023toolllm, zeng2023agenttuning}. {\framework} builds upon the instruction-following paradigm and focuses on tool-using ability through the disentanglement of CoS-Planning and CoS-Calling, which makes learning more efficient.

\section{Toolink Framework}

\begin{figure*}[!t]
    \centering
    \includegraphics[width=\linewidth]{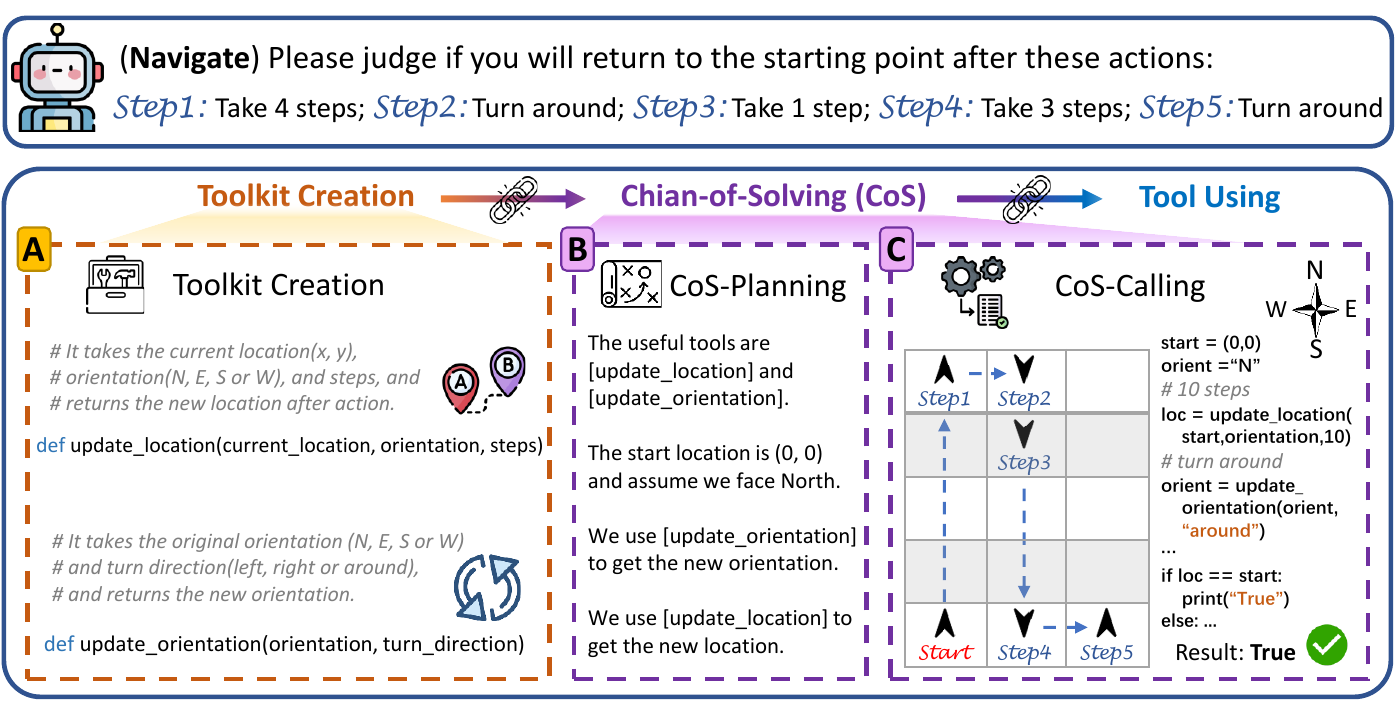}
    \caption{A problem solving chain of Toolink pipeline. We show an example from task Navigate. Toolink first creates a toolkit generally applicable to the task, and then approaches the specific query through CoS, which involves planning and calling of the created tools.}
    \label{fig:main_pipeline}
\end{figure*}

As shown in \Cref{fig:main_pipeline}, Toolink first adopts toolkit creation to break down the target task through generating potential tools for task-solving (\Cref{method:creation}). Then, the model links these created tools to address specific queries by selecting pertinent tools from the toolkit, planning their uses, and performing tool invocations (\Cref{method:cos}). This new reasoning approach, referred to as chain-of-solving (CoS), not only enables the effective and coherent application of tools but also facilitates the tool-using adaptation on the open-source model (\Cref{LLaMA_Adaptation}).




\subsection{Toolkit Creation}
\label{method:creation}
Toolkit creation decomposes a general task into modular and essential tools for problem-solving, facilitating more flexible tool utilization.

\paragraph{Overview.} Given the target task $T$, toolkit creation breaks it down into more manageable components $t_1, t_2, ..., t_n$ through generating a toolkit $K_T = \{k_1, k_2, ..., k_n\}$, where $k_i (i \leq n)$ represents the tool to solve the subtask $t_i$. We illustrate our approach in \Cref{fig:main_pipeline}A, where the target task $T=\texttt{Navigate}$ is decomposed into $t_1$ (movement in a single direction) and $t_2$ (change of orientation). Each component is represented by a specific implementation encapsulated within a function tool.

\paragraph{Toolkit Making.} We utilize ChatGPT for task decomposition. For each task $T$, we provide ChatGPT with a task description and a few data samples $D_{T\texttt{-sample}}$, expecting them to facilitate the model's understanding of task $T$'s objective and identify commonalities among queries. The prompting details are presented in \Cref{Toolkit_Creation} and \Cref{fig:apdx_prompt}. Note that our design requires only a few data points as demonstrations fed into the closed-source ChatGPT, leaving the entire testing set for local processing to maintain privacy.

\paragraph{Tool Details.} Each tool $k_i$ within the toolkit $K_T$ is comprised of a concise introduction and its corresponding code implementation. The introduction provides a brief overview of $k_i$'s utility, inputs, and outputs, facilitating effective planning and calling in subsequent steps.



\subsection{Chain-of-Solving}
\label{method:cos}
Chain-of-solving (CoS) involves deliberate planning and decision-making for tool invocation, which bridges the gap between toolkit creation and downstream tool use for task query resolution. CoS is disentangled into CoS-Planning and CoS-Calling. This separation allows for a more transparent and interpretable reasoning path, thereby enhancing the applicability of CoS to open-source models.

\paragraph{CoS-Planing.} The CoS-Planning stage entails selecting useful tools from a toolkit $K_T$ in response to a specific query of task $T$. It employs natural language-based reasoning chains, referred to as a \emph{plan}, to determine the most effective way to utilize the selected tools to solve the given query.

In \Cref{fig:main_pipeline}B, the model devises strategies for employing tools to update the location and orientation, with additional initial conditions that serve as a guiding hint. Planning plays a crucial role in establishing a link between toolkit creation and decision-making, thus reducing the cognitive burden associated with tool-use reasoning.



\paragraph{CoS-Calling.} The CoS-Calling stage entails the utilization of selected tools and interpretation of the plan into program language to perform tool calls. The plan generated in the previous stage serves as the guidance for program implementation. During the tool execution, all results from the tool invocations are implicitly captured and used to extract the final answer for the given query.

\Cref{fig:main_pipeline}C illustrates this process, where the model simulates the entire navigation process using code as the underlying medium. In this example, the model derives the final correct answer, thereby demonstrating a successful CoS-Calling process.



\begin{table}[!t]
\centering
\small
\tabcolsep=0.015\linewidth
\begin{tabular*}{\linewidth}{cccc}
\toprule
\textbf{Category} & \textbf{Set Name} & \textbf{\makecell{Source}} & \textbf{\makecell{Number}} \\
\midrule
\multirow{2}{*}{\textbf{\makecell{Tool-Using}}}
& Tool-Planning & Augmented & 4.4K \\
& Tool-Calling & Augmented & 4.4K \\
\midrule
\multirow{6}{*}{\textbf{\makecell{Code\\Generation}}}
& Python-Simple & New & 2.0K \\
& Python-Specific & New & 2.0K \\
& Math & Augmented & 2.5K \\
& Algorithm & Github & 2.3K \\
& LeetCode & LeetCode & 0.8K \\
& Rectification & Sources Above & 1.6K \\
\midrule
\multirow{1}{*}{\textbf{\makecell{Total}}}
& - & - & 20.0K \\
\bottomrule
\end{tabular*}
\caption{The statistics about the sources and number of data points in each category of CoS-GPT. \textit{Augmented} represents augmented from an existing dataset.}
\label{tab:Base_Training_Stat}%
\end{table}


\subsection{Open-Source Model Adaptation}
\label{LLaMA_Adaptation}
Considering the limited adaptability, high inference cost, and privacy concerns posed by closed-source models, we aim to enhance the CoS ability in open-source models. We propose the CoS-GPT, a specialized training dataset that emphasizes the planning and calling of tools, along with code generation. These elements are crucial for boosting the model's CoS ability. The statistics related to CoS-GPT are presented in \Cref{tab:Base_Training_Stat}. Furthermore, for each specific target task $T$, we employ $D_{T\texttt{-sample}}$ to generate a task-specific dataset. This is achieved by augmenting each sample query with suitable tools, thereby facilitating a more effective training of task $T$ on the open-source models.

\paragraph{Construction of CoS-GPT.} To enhance the open-source model's skills in applying tools for problem-solving, we construct CoS-GPT from scratch to improve the model's CoS ability from planning, calling, and coding. We include the first two aspects as they are essential for CoS within Toolink, and the last aspect as it serves as the medium for tool-using.

For data points about planning and calling, we enhance the AQUA-RAT~\citep{ling2017program}, GSM8K~\citep{cobbe2021training}, and TabMWP~\citep{lu2022dynamic} datasets, comprising graduate-level math problems, numerical reasoning tasks, and diverse table contents, with toolkits. Each query is augmented with a toolkit containing both useful and redundant tools. The model's objective for planning is to select and plan the use of useful tools, while for calling, the objective is to learn how to call the chosen tools through codes. We apply ChatGPT to simulate this process and utilize their responses for dataset construction. Please refer to \Cref{Tool-Using_Data_Details} for more details.

Data points for code generation encompass diverse sources, including augmentation from existing datasets, GitHub repositories, and newly generated data, detailed in \Cref{Code_Generation_Details}. Each query adheres to an instruction-following pattern and aims to enhance the open-source model's understanding of code while expanding its versatility in making informed decisions when performing CoS.


\begin{table}[!t]
\centering
\small
\tabcolsep=0.02\linewidth
\begin{tabular*}{\linewidth}{cc}
\toprule
\textbf{Category} & \textbf{Task Name} \\
\midrule
\multirow{2}{*}{\textbf{\makecell{Mathematics}}}
& Arithmetic, Matrix Shape, \\
& Chinese Remainder \\
\midrule
\multirow{1}{*}{\textbf{\makecell{Common Sense}}}
& Date Understanding, Navigate \\
\midrule
\multirow{1}{*}{\textbf{\makecell{Logical Reasoning}}}
& Dyck Language, Boolean Expression \\
\midrule
\multirow{1}{*}{\textbf{\makecell{Decomposition}}}
& Tracking Shuffled Objects \\
\bottomrule
\end{tabular*}
\caption{The categories of 8 BIG-bench tasks tested.}
\label{tab:task_category}%
\end{table}


\paragraph{Construction of Task-Specific Data.} For each target task $T$, we construct 200 tool-augmented data points (100 each for plan and call) from the publicly available samples $D_{T\texttt{-sample}}$, and use them to tune the open-source model together with CoS-GPT. Similar to the construction process for tool-using data in CoS-GPT, we first augment $T$ with a toolkit $K_{T}$. Next, we employ ChatGPT to select useful tools for each query and generate the calling decision. The decision's output is compared against the standard answer, and minor adjustments may be made to ensure the augmented data's validity.

\paragraph{Open-Source Model Finetuning.} Together with CoS-GPT, we apply the tool-augmented data points from all target tasks to finetune the open-source model. We expect the derived tool-augmented open-source model to excel in applying useful tools for problem-solving. By planning and calling through CoS, this model links the created toolkit with specific queries, which realizes the final goal of the Toolink framework.


\begin{table*}[!t]
    \centering
    \small
    \tabcolsep=0.0095\linewidth
    \begin{tabular*}{\linewidth}{ccccccccccc}
    \toprule
    \textbf{Task} & \textbf{Arith.} & \textbf{\makecell{Date U.}} & \textbf{\makecell{Matrix S.}} & \textbf{\makecell{Navigate}} & \textbf{\makecell{Chinese R.}} & \textbf{\makecell{Dyck L.}} & \textbf{\makecell{Boolean E.}} & \textbf{\makecell{Tracking S.}} & \textbf{\makecell{Average}} \\
    \midrule
    \multirow{1}{*}{\textbf{\makecell{Num. of Tools}}}
    & 5 & 3 & 5 & 2 & 2 & 4 & 2 & 4 & 3.38 \\
    \midrule
    \midrule
    \multirow{1}{*}{\textbf{\makecell{Vanilla}}}
    & 77.78 & 68.67 & 40.90 & 65.16 & 0.0 & 19.40 & 80.70 & 23.67 & 47.03 \\
    \midrule
    \multirow{1}{*}{\textbf{\makecell{CoT}}}
    & 79.44 & 68.67 & 80.46 & 87.96 & 0.0 & 19.42 & 75.88 & 40.78 & 56.58 \\
    \midrule
    \multirow{1}{*}{\textbf{\makecell{CoS}}}
    & 100.00 & 69.28 & 93.67 & 85.30 & 95.14 & 52.46 & 97.37 & 99.11 & \textbf{86.54} \\
    \midrule
    \midrule
    \multirow{1}{*}{\textbf{\makecell{CoS-Planning}}} 
    & 100.00 & 66.16 & 95.18 & 94.78 & 100.00 & 74.58 & 95.39 & 99.85 & 90.74 \\
    \midrule
    \midrule
    \multirow{1}{*}{\textbf{\makecell{CoS-Calling}}} 
    & 100.00 & 90.96 & 97.44 & 88.44 & 95.67 & 98.55 & 93.42 & 100.00 & 95.56 \\
    \bottomrule
    \end{tabular*}
    \caption{We record the number of tools in the toolkit created for each task and demonstrate the accuracy (\%) of ChatGPT on 8 BIG-bench tasks under different settings. We report the results of Vanilla, CoT baselines, and our CoS method, and report the performance of CoS-Planning and CoS-Calling separately.}
    \label{tab:ChatGPT_result}%
\end{table*}
    
    
\begin{table*}[!t]
    \centering
    \small
    \tabcolsep=0.0085\linewidth
    \begin{tabular*}{\linewidth}{ccccccccccc}
    \toprule
    \textbf{Method} & \textbf{Model} & \textbf{Arith.} & \textbf{\makecell{Date U.}} & \textbf{\makecell{Matrix S.}} & \textbf{\makecell{Navigate}} & \textbf{\makecell{Chinese R.}} & \textbf{\makecell{Dyck L.}} & \textbf{\makecell{Boolean E.}} & \textbf{\makecell{Tracking S.}} \\
    \midrule
    \multirow{3}{*}{\makecell{\textbf{CoT}\\(\textit{Prompting} \\ \textit{w/ demo})}}
    & Alpaca & 19.89 & 39.76 & 5.62 & 47.11 & 0.0 & 0.0 & 57.46 & 0.44 \\
    & LLaMA-7B & 39.44 & 33.73 & 12.58 & 39.70 & 0.0 & 2.90 & 50.44 & 14.22 \\
    & ChatGPT & 79.44 & 68.67 & 80.46 & 87.96 & 0.0 & 19.42 & 75.88 & 40.78 \\
    \midrule
    \multirow{1}{*}{\textbf{\makecell{CoT}} (\textit{Tuned})}   
    & LLaMA-CoT & 50.44 & 49.40 & 70.82 & 71.64 & 0.0 & 35.27 & 62.72 & 28.44 \\
    \midrule
    \multirow{3}{*}{\makecell{\textbf{CoS}\\(\textit{Prompting} \\ \textit{w/ demo})}}  
    & Alpaca & 17.78 & 7.83 & 3.00 & 48.60 & 7.56 & 1.00 & 94.74 & 6.78 \\
    & LLaMA-7B & 55.89 & 17.47 & 10.65 & 45.90 & 23.80 & 35.83 & 99.12 & 0.67 \\
    & ChatGPT & \textbf{100.00} & 69.28 & \textbf{93.67} & 85.30 & 95.14 & 52.46 & 97.37 & 99.11 \\
    \midrule
    \multirow{1}{*}{\textbf{\makecell{CoS}} (\textit{Tuned})}
    & LLaMA-CoS & \textbf{100.00} & \textbf{74.10} & 91.01 & \textbf{99.56} & \textbf{95.44} & \textbf{98.21} & \textbf{100.00} & \textbf{99.56} \\
    \bottomrule
    \end{tabular*}
    \caption{The accuracy (\%) of baselines and LLaMA-CoS on the 8 BIG-bench tasks. LLaMA-CoS employs planning and calling of tools, which beats all CoT baselines by large margins and is on par with ChatGPT's CoS ability.}
    \label{tab:main_result}%
\end{table*}
    
\section{Experiments}
To evaluate the effectiveness of {\framework}, we first conduct a validation test using ChatGPT. We select eight distinct tasks from the BIG-bench dataset~\citep{srivastava2022imitation} to investigate whether {\framework} can effectively leverage ChatGPT's creativity and tool-using capability to improve task performance.

Subsequently, we finetune the open-source LLaMA-7B model following the adaptation process outlined in \Cref{LLaMA_Adaptation}. This results in LLaMA-CoS, a model capable of linking the created toolkit with specific tool use through CoS. We evaluate the effectiveness of LLaMA-CoS in utilizing tools on the same set of eight tasks and showcase its excellence.

\subsection{Validation Evaluation}
\label{Validation_Test}
\paragraph{Settings.} To evaluate the effectiveness of {\framework}, we conducted a validation test using ChatGPT on eight tasks of diverse categories from BIG-bench, as detailed in \Cref{tab:task_category}. For each task, we first employ ChatGPT to create a toolkit, outlined in \Cref{method:creation}. The total number of tools in the toolkit created for each task is shown in \Cref{tab:ChatGPT_result}, with the tool's implementation details provided in \Cref{Task_Tools}. Equipped with these tools, ChatGPT is presented with instructions and demonstration examples to perform CoS for problem-solving, detailed in \Cref{Composition_Stage_Settings}.

\paragraph{Baselines.} We compare CoS against two baselines: i) \textbf{Vanilla} baseline, where ChatGPT directly produces the final answer. ii) \textbf{CoT} baseline~\citep{weichain}, where ChatGPT performs chain-of-thought reasoning before providing an answer.

\paragraph{Evaluations.} We evaluate the ability of ChatGPT to leverage plans and calls to perform CoS. The accuracy is measured by matching the model's final output to the correct answer. In addition, we also evaluated the individual contributions of CoS-Planning and CoS-Calling separately.

During CoS-Planning, the model is asked to select useful tools and plan their uses given the created toolkit. The planning accuracy is thus measured by the following metric:
\begin{equation}
\begin{aligned}
\label{eq_acc_toolplan}
ACC = \max \{ \frac{|K_{\texttt{correct}}| - |K_{\texttt{error}}|}{|K_{\texttt{correct}}| + |K_{\texttt{error}}|} , 0\},
\end{aligned}
\end{equation}
where $|K_{\texttt{correct}}|$ denotes the number of correct (useful) tools in the toolkit selected in the model's generated plan, while $|K_{\texttt{error}}|$ denotes the number of incorrect (redundant) tools selected.

During CoS-Calling, the model is asked to implement the plan using code as the medium, after the useful tools are selected. The accuracy is thus measured by matching the output from the final execution with the correct answer. Please refer to \Cref{Separation_Plan_Call} for more details regarding the separated test of CoS-Planning and CoS-Calling.

\paragraph{Results.} The results are presented in \Cref{tab:ChatGPT_result}. ChatGPT which uses tools through the CoS approach achieves significantly improved performance compared to other baselines, with notable margins of superiority. Further, the accuracy for CoS-Calling and CoS-Planning individually is even higher, indicating successful reasoning in each step of CoS which links toolkit creation with specific uses. These findings affirm the validity of {\framework}, establishing a strong basis for its potential transferability to smaller, open-sourced models.

\subsection{Experiments on LLaMA-CoS}
Our primary objective is to apply {\framework} to smaller, open-sourced models. To this end, the models from the LLaMA family~\citep{touvron2023llama} stand out due to their capability to perform reasoning and generate codes. Considering the affordability of computational resources, we select LLaMA-7B as the representative base model to evaluate the performance of {\framework} on open-source models.

\paragraph{Obtaining LLaMA-CoS.} We follow the adaptation process outlined in \Cref{LLaMA_Adaptation} and finetune LLaMA-7B with CoS-GPT and eight sets of task-specific tool-augmented data. The eight tasks are the same ones as those we test in \Cref{Validation_Test}. Applying the training setting detailed in \Cref{Exp_Setting_Details}, we derive a powerful variant, LLaMA-CoS, that excels in using tools through CoS.

\paragraph{Settings.} We use LLaMA-CoS as the representative finetuned open-source model for testing. Building upon the validation test conducted on ChatGPT, we evaluate LLaMA-CoS's performance on the same set of eight tasks from BIG-bench. We keep all the metrics the same as in \Cref{Validation_Test}

\paragraph{Baselines.} As a comparison to CoS, we employ the chain-of-thought (CoT) reasoning as the baseline. We evaluate both methods under two scenarios: i) Prompting with demonstrations on Alpaca, LLaMA-7B, and ChatGPT, and ii) Finetuning specifically on the LLaMA-7B model. We referred to the LLaMA-7B tuned with CoT data as LLaMA-CoT, while our model, LLaMA-CoS, is specially tuned for tool use through CoS.

\paragraph{Results.} We present the results in \Cref{tab:main_result}. Notably, LLaMA-CoS achieves an impressive average accuracy of 94.74\%, outperforming all the CoT baselines, whether tuned or not, by a substantial margin. Compared to ChatGPT, which exhibits strong reasoning and tool-using capabilities under the CoS setting, our tuned model can still achieve comparable performance. These results highlight the effectiveness of CoS in outperforming traditional CoT methods and demonstrate the successful transfer of tool-using abilities from closed-source LLMs to smaller, open-source models.


\begin{table*}[!t]
\centering
\small
\tabcolsep=0.0075\linewidth
\begin{tabular*}{\linewidth}{cccccccccc}
\toprule
\textbf{Method} & \textbf{Model} & \textbf{Arith.} & \textbf{\makecell{Date U.}} & \textbf{\makecell{Matrix S.}} & \textbf{\makecell{Navigate}} & \textbf{\makecell{Chinese R.}} & \textbf{\makecell{Dyck L.}} & \textbf{\makecell{Boolean E.}} & \textbf{\makecell{Tracking S.}} \\
\midrule
\multirow{1}{*}{\textbf{\makecell{CoS-Whole}}}
& LLaMA-CoS & \textbf{100.00} & 74.10 & 91.01 & 99.56 & 95.44 & 98.21 & 100.00 & 99.56 \\
\midrule
\midrule
\multirow{3}{*}{\makecell{\textbf{CoS-Planning}\\(\textit{Prompting} \\ \textit{w/ demo})}}
& Alpaca & 18.22 & 27.41 & 24.15 & 77.16 & \textbf{100.00} & 76.3 & \textbf{97.59} & 99.37 \\
& LLaMA-7B & 74.11 & 27.71 & 25.02 & 77.16 & \textbf{100.00} & 93.80 & \textbf{97.59} & \textbf{100.00} \\
& ChatGPT & \textbf{100.00} & 66.16 & \textbf{95.18} & 94.78 & \textbf{100.00} & 74.58 & 95.39 & 99.85 \\
\midrule
\multirow{1}{*}{\textbf{\makecell{CoS-Planning}}}
& LLaMA-CoS & \textbf{100.00} & \textbf{85.84} & 89.62 & \textbf{97.14} & \textbf{100.00} & \textbf{99.19} & \textbf{97.59} & \textbf{100.00} \\
\midrule
\midrule
\multirow{3}{*}{\makecell{\textbf{CoS-Calling}\\(\textit{Prompting} \\ \textit{w/ demo})}} 
& Alpaca & 99.44 & 24.70 & 30.08 & 48.60 & 17.97 & 1.56 & 89.91 & 6.78 \\
& LLaMA-7B & 74.70 & 51.20 & 55.49 & 43.77 & 24.81 & 25.67 & 94.30 & 1.56 \\
& ChatGPT & \textbf{100.00} & 90.96 & \textbf{97.44} & 88.44 & \textbf{95.67} & \textbf{98.55} & 93.42 & \textbf{100.00} \\
\midrule
\multirow{1}{*}{\textbf{\makecell{CoS-Calling}}} 
& LLaMA-CoS & \textbf{100.00} & \textbf{91.57} & 95.56 & \textbf{98.88} & 94.18 & \textbf{98.55} & \textbf{95.61} & 88.44 \\

\bottomrule
\end{tabular*}
\caption{The accuracy (\%) of CoS-Planning and CoS-Calling separately on 8 BIG-bench tasks. Results show LLaMA-CoS has excellent ability in understanding and using tools through CoS.}
\label{tab:main_plancall_results}%
\end{table*}


\subsection{Results Analysis}
\label{Results_Analysis}
\paragraph{Excellence in Both Planning and Calling.} 
To comprehensively assess the CoS method, we similarly report the individual contribution of CoS-Planning and CoS-Calling in \Cref{tab:main_plancall_results}. Our results demonstrate that CoS-Planning and CoS-Calling separately surpass the performance achieved by CoT-based models on all tasks. This validates the model's proficiency in both stages during CoS and underscores the rationale behind designing CoS-Planning and CoS-Calling to promote effective tool use under the {\framework} framework.


\begin{figure}[t]
    \centering
    \includegraphics[width=0.95\linewidth]{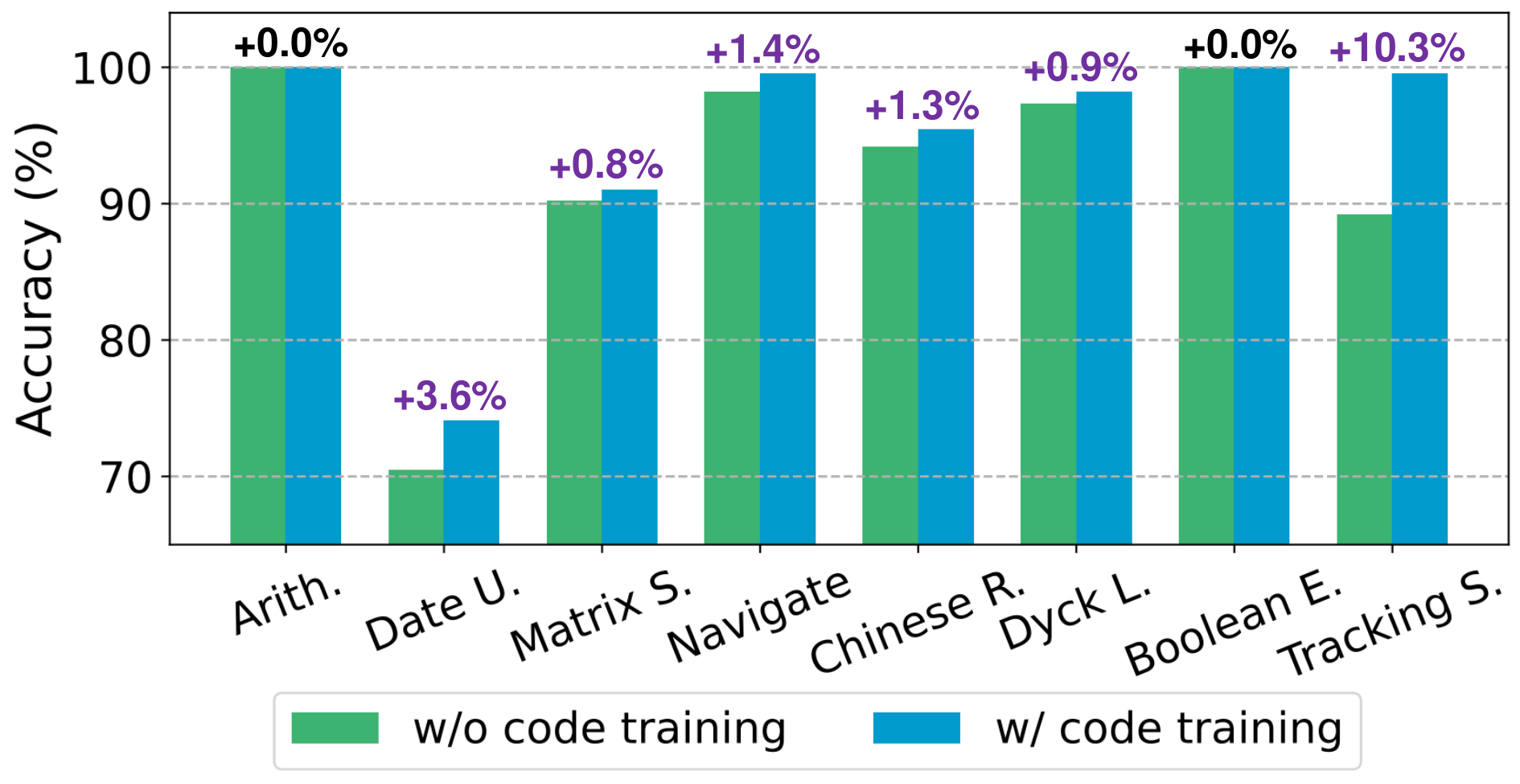}
    \caption{The improvement of performance when code generation data points are involved during training.}
    \label{fig:code_ablation}
\end{figure}


\paragraph{Necessity of Code Training.}
To evaluate the impact of code generation data in CoS-GPT, we compare the LLaMA-7B tuned with or without them. The results in \Cref{fig:code_ablation} indicate that LLaMA-CoS trained with code generation data achieves higher accuracy, with an average improvement of 1.4\%. This validates the necessity of training on code generation together with CoS ability. By incorporating these data points, the model learns to leverage codes as the medium for tool-using more effectively, which ultimately enhances task performance.


\begin{figure}[t]
    \centering
    \includegraphics[width=\linewidth]{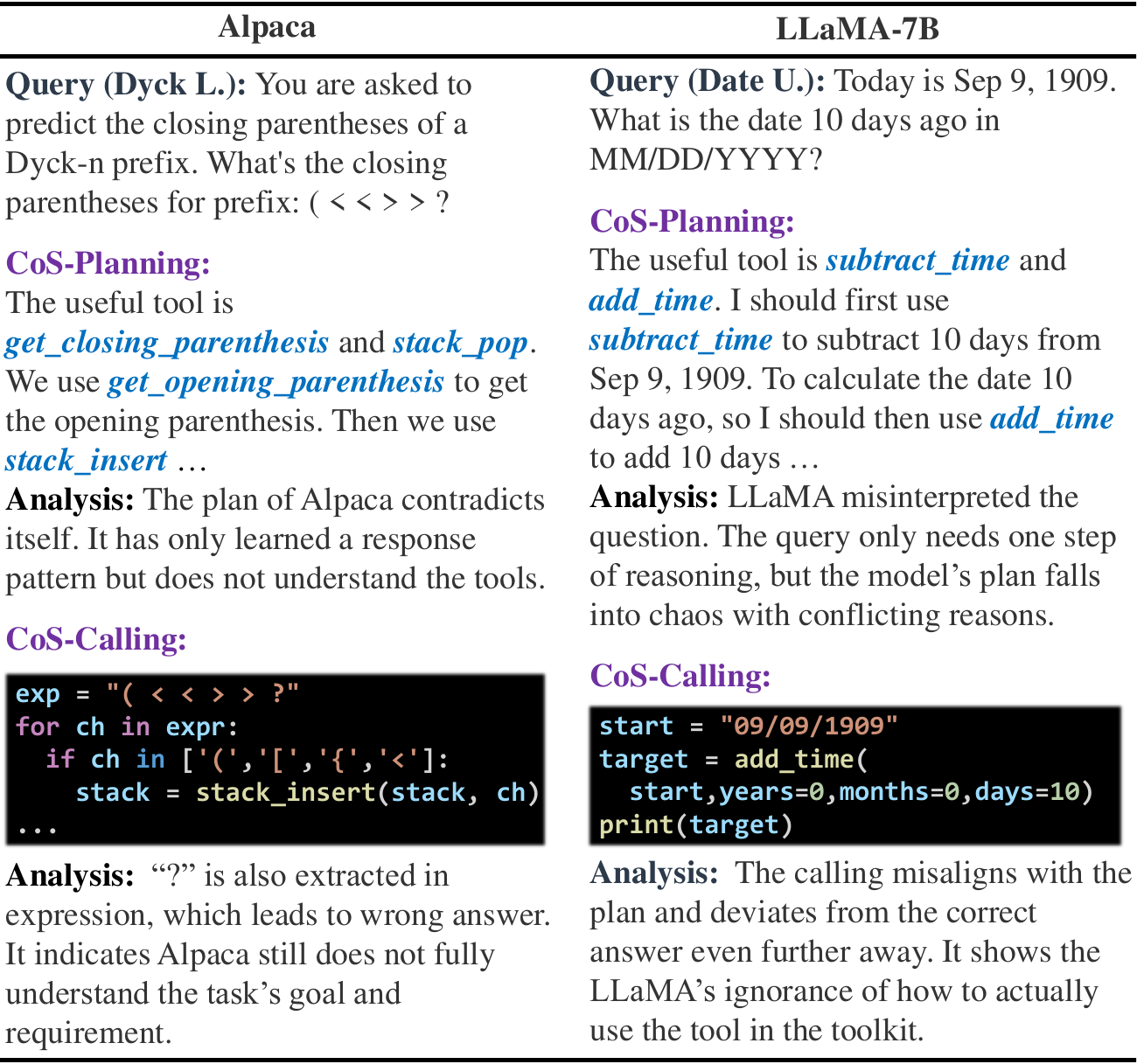}
    \caption{Detailed error analysis of Alpaca and LLaMA-7B regarding CoS-Planning and CoS-Calling.}
    \label{fig:error_analysis}
\end{figure}


\paragraph{Error Analysis of LLaMA-7B and Alpaca.}
We discover from the results that the raw LLaMA-7B and Alpaca's performance lags far behind. To provide insights into why they fail to do CoS-Planning and CoS-Calling even with demonstrations, we conduct a detailed error analysis in \cref{fig:error_analysis}. 

Upon analyzing the errors made by both models, we identified two primary tendencies: i) the models tend to learn the pattern but often overlook the utility of the tool and the purpose of the task; ii) they frequently exhibit disarray in reasoning and a misalignment between the tool plan and the tool call. These issues contribute significantly to the subpar performance of both models.

\begin{figure*}[!t]
\centering
\subfigure[Sequential Tool Calling.] { \label{fig:apdx_case:sequential} 
\includegraphics[width=\linewidth]{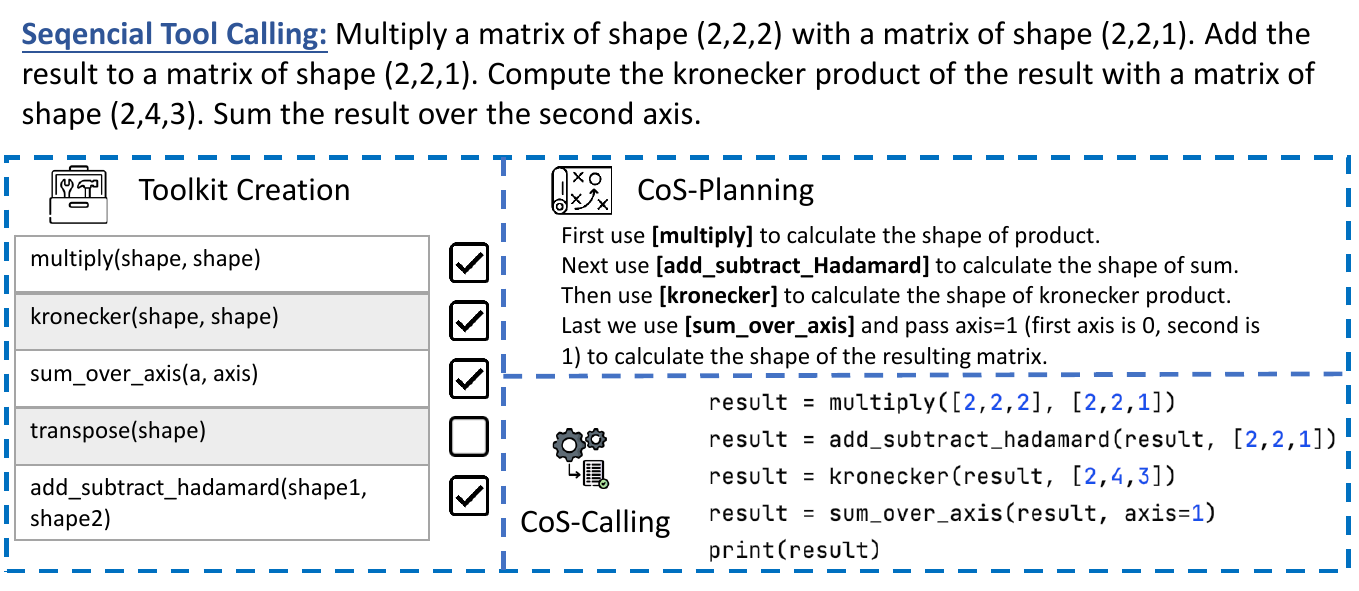}}
\subfigure[Conditional Tool Calling.] { \label{fig:apdx_case:conditional} 
\includegraphics[width=\linewidth]{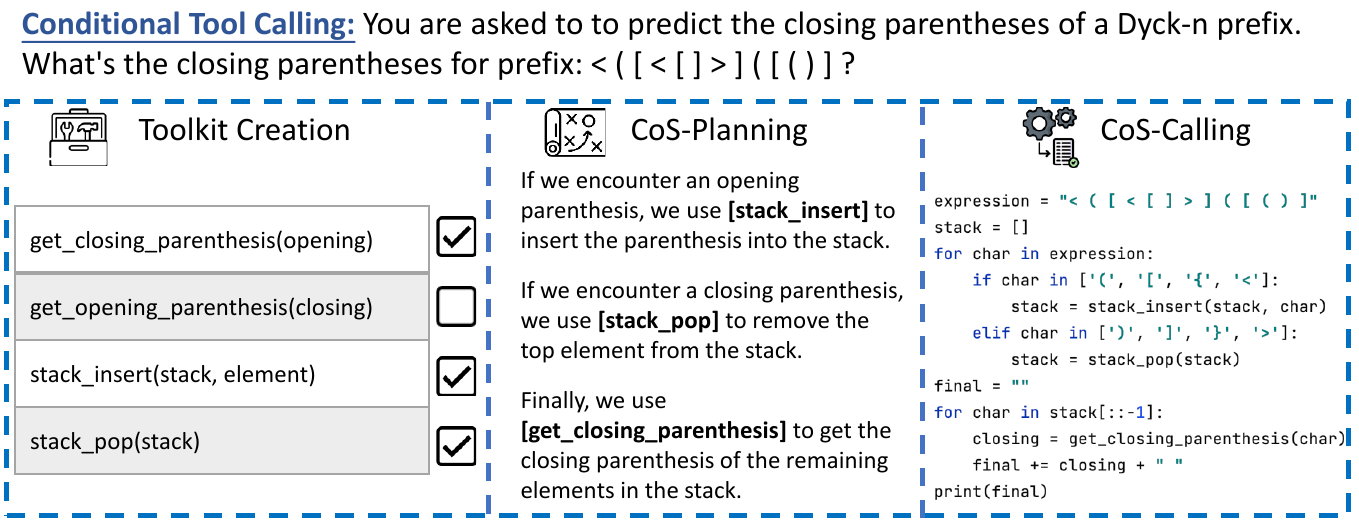}}
\subfigure[Nested Tool Calling.] { \label{fig:apdx_case:nested} 
\includegraphics[width=\linewidth]{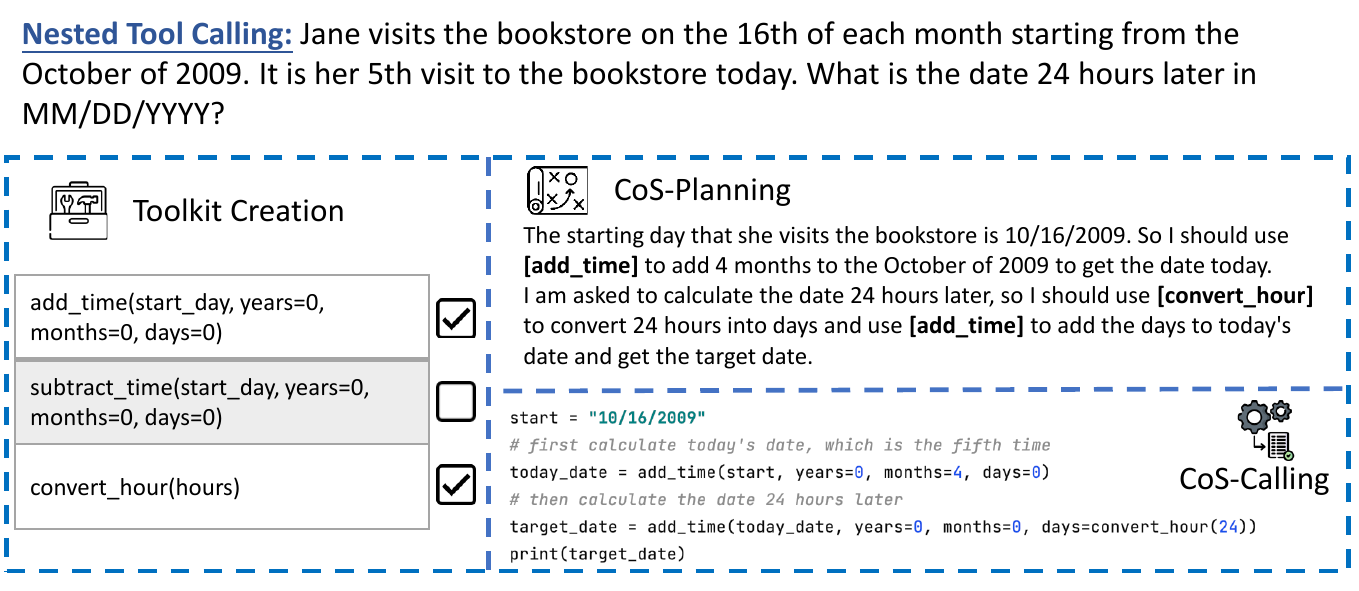}}
\caption{Case Studies on the diverse CoS-Calling patterns in the main experiment.}
\label{fig:apdx_case}
\end{figure*}

\paragraph{Diverse Usage of Toolkit.}
Throughout the experiments, LLaMA-CoS exhibits diverse CoS-Calling patterns. It is capable of sequentially calling different tools to achieve a specific purpose, using tools in a non-linear logic (in a loop or with conditions), or performing nested tool calls, where the output from one tool directly serves as the other tool's input. These abilities illustrate the robustness and adaptability of LLaMA-CoS across diverse scenarios. We provide case studies and more details in \Cref{diverse_cos_pattern} and \Cref{fig:apdx_case}.

\section{Further Studies}

\begin{table}[!t]
\centering
\small
\tabcolsep=0.015\linewidth
\begin{tabular*}{\linewidth}{ccccc}
\toprule
\textbf{Task} & \textbf{\makecell{If in\\CoS-GPT}} & \textbf{\makecell{CoS\\Stage}} & \textbf{\makecell{LLaMA\\-CoS}} & \textbf{\makecell{ChatGPT}} \\
\midrule
\multirow{2}{*}{\textbf{\makecell{AQUA-RAT}}} & \multirow{2}{*}{\green{\usym{2713}}} & \textit{Planning} & \textbf{58.80} & 52.90 \\
   &   & \textit{Calling} & 56.12 & \textbf{65.94} \\
\midrule
\multirow{2}{*}{\textbf{\makecell{MATH}}} & \multirow{2}{*}{\green{\usym{2713}}} & \textit{Planning} & \textbf{65.83} & 52.61 \\
   &   & \textit{Calling} & \textbf{50.75} & 43.25 \\
\midrule
\multirow{2}{*}{\textbf{\makecell{TabMWP}}} & \multirow{2}{*}{\green{\usym{2713}}} & \textit{Planning} & \textbf{90.00} & 60.75 \\
   &   & \textit{Calling} & \textbf{66.00} & 32.75 \\
\midrule
\multirow{2}{*}{\textbf{\makecell{FinQA}}} & \multirow{2}{*}{\red{\usym{2717}}} & \textit{Planning} & \textbf{70.51} & 50.15 \\
   &   & \textit{Calling} & 22.38 & \textbf{32.38} \\
\midrule
\multirow{2}{*}{\textbf{\makecell{GSM8K}}} & \multirow{2}{*}{\red{\usym{2717}}} & \textit{Planning} & \textbf{61.29} & 53.83 \\
   &   & \textit{Calling} & \textbf{57.25} & 36.50 \\
\bottomrule
\end{tabular*}
\caption{The accuracy (\%) of CoS-Planning and CoS-Calling on five diverse datasets applying LLaMA-CoS or ChatGPT. Results show that LLaMA-CoS generally beats ChatGPT and is robust to unseen tasks.}
\label{tab:generalize_unseen_task}%
\end{table}


\begin{table}[!t]
\centering
\small
\tabcolsep=0.02\linewidth
\begin{tabular*}{\linewidth}{cccc}
\toprule
\textbf{Task} & \textbf{\makecell{Toolkit Origin}} & \textbf{\makecell{LLaMA-CoS}} & \textbf{\makecell{ChatGPT}} \\
\midrule
\multirow{2}{*}{\textbf{\makecell{Dynamic\\Cnt.}}}   
& \textit{Raw} & 97.50 & 80.83 \\
& \textit{From Dyck L.} & 73.30 & 79.17 \\
\midrule
\multirow{2}{*}{\textbf{\makecell{Unit\\Interp.}}} 
& \textit{Raw} & 70.83 & 80.83 \\
& \textit{From Arith.} & 65.83 & 80.00 \\
\bottomrule
\end{tabular*}
\caption{The accuracy (\%) of ChatGPT and LLaMA-CoS, with toolkit newly created for the target task (\textit{Raw}) or borrowed from other tasks. Our results show that both ChatGPT and LLaMA-CoS can utilize tools not specifically tailored for the target task through CoS.}
\label{tab:generalization_toolkit}%
\end{table}


In this section, we show the generalization of LLaMA-CoS to novel tasks and how it can use toolkits that are not specially tailored for solving the target task. These studies aim to illustrate the robustness of LLaMA-CoS in utilizing tools through planning and calling.

\subsection{Generalization to Novel Tasks}
The eight evaluation tasks~\citep{srivastava2022imitation} we previously used all have data points presented during training, despite only leveraging a few publicly available samples. To showcase the generalization ability of LLaMA-CoS, we further test it on two new tasks: FinQA~\citep{chen2022finqa} and GSM8K~\citep{cobbe2021training}. FinQA involves questions based on financial report data, while GSM8K involves grade school math problems. 

Together with AQUA-RAT, MATH, and TabMWP, whose data are presented in CoS-GPT (as detailed in \Cref{LLaMA_Adaptation}), we randomly select a maximum of 400 test data points from each of the five tasks, ensuring they \textit{do not} appear in CoS-GPT. We augment each data point with a toolkit, following the method outlined in \Cref{LLaMA_Adaptation} regarding how CoS-GPT is constructed. In experiments, we follow the CoS-Planning and CoS-Calling tests outlined in \Cref{Validation_Test}.

We show in \Cref{tab:generalize_unseen_task} that LLaMA-CoS achieves high accuracy in both planning and calling stages and could even beat ChatGPT in performance. This affirms the effectiveness and robustness of its CoS ability even applied to unseen tasks. As our tests encompass math, finance, table reasoning, etc, this finding also emphasizes \textbf{the robustness of LLaMA-CoS across diverse types of tasks}.

\subsection{CoS on Generic Toolkit}
We further explore the ability of LLaMA-CoS to use a generic toolkit instead of the one specifically tailored for the target task. In real-world scenarios, toolkits are usually designed to address diverse tasks rather than tailored for a single task. We expect LLaMA-CoS and ChatGPT can apply toolkits borrowed from other tasks to solve the target query in a CoS approach.

To validate this, we source two additional tasks from BIG-bench: Dynamic Counting and Unit Interpretation. For each task, we provide a toolkit created explicitly for the target task or borrowed from another task. Specifically, we pair Dynamic Counting and Unit Interpretation respectively with Dyck Language and Arithmetic. 

Under these settings, we evaluate the performance of both LLaMA-CoS and ChatGPT in \Cref{tab:generalization_toolkit} and show that \textbf{both LLaMA-CoS and ChatGPT can utilize a generic toolkit borrowed from another task to solve target queries through CoS}. Though the performance still lags, these findings nevertheless confirm our assumption that CoS can increase the robustness of tool-using, and make our {\framework} more applicable to real-world scenarios. We present more details in \Cref{Toolkit_Generalization_Details}.

\section {Conclusions}
We present {\framework}, a tool-training framework that effectively applies toolkits to solve problems leveraging small, open-source language models. {\framework} offers increased flexibility in adapting to diverse downstream tasks while addressing concerns related to high inference costs and privacy. Our main contributions include i) empirically implementing a framework that can effectively leverage open-source models' tool-using ability, ii) devising the chain-of-solving (CoS) method that links toolkit creation and tool use through robust planning and calling, and iii) releasing the CoS-GPT dataset that aims to enhance the model's CoS capabilities. 

Specifically, our LLaMA-CoS model outperforms traditional CoT and achieves a comparable performance to ChatGPT concerning tool-using. We believe our study provides a solid foundation for future researchers to explore and enhance the tool-using capabilities of open-source models.

\section*{Limitations}
Our experiments focus on equipping the open-source model with tool-using capabilities through the CoS approach, specifically in planning and calling, while excluding the ability to create toolkits. This limitation arises from the fact that the LLaMA-7B primarily relies on provided demonstrations and lacks the internal creativity required for toolkit creation. Moreover, the absence of enough training data further hampers the acquisition of this knowledge. We acknowledge this challenge posed by the transfer of the toolkit creation capability from closed-source models and leave it as an avenue for future research.

Additionally, it is important to note that though the tasks tested in our study include diverse toolkits and queries, they are mostly sourced from the BIG-bench dataset. To gain a more holistic understanding of the generalizability of our results, future research should expand the application of {\framework} to a broader range of scenarios. This would enable a more comprehensive assessment of the framework's efficacy and applicability.

\section*{Ethics Statement}
We consider the following issues in this paper:
\begin{itemize} [topsep=1pt, partopsep=1pt, leftmargin=12pt, itemsep=-3pt]
    \item \textbf{Privacy} is a crucial aspect to consider when utilizing closed-source models such as ChatGPT and GPT4. These models have the potential to learn sensitive information internally, posing a risk to personal privacy. In contrast, {\framework} addresses this concern by leveraging only a limited number of publicly available samples for toolkit creation, leaving the majority of testing queries blind to closed-source LLMs. This approach reduces the possibility of mishandling data and safeguards user privacy. By minimizing the exposure of sensitive information, {\framework} mitigates the risks associated with privacy breaches when compared to closed-source models.
    \item \textbf{Transparency} is a key aspect that aims to enhance the interpretability and comprehensibility of AI systems from a human perspective. In {\framework}, we prioritize transparency through the creation of toolkits that provide clear information about their utility, inputs, and outputs. Additionally, we disentangle the CoS into separate stages of planning and calling, which increases the interpretability of the model's reasoning for users. We also encourage future research to further document the specific scenarios in which our framework exhibits its maximum effectiveness, as well as to outline potential risks involved. This will contribute to a more comprehensive understanding of {\framework} and facilitate informed decision-making.
    \item \textbf{Potential Bias} is another critical aspect that we prioritize addressing in our work. We acknowledge that bias and discrimination can inadvertently manifest through problematic examples present in the training data. To mitigate this concern, we adopt a meticulous approach to curate the CoS-GPT dataset, which consists of data points from various sources. We emphasize diversity to minimize the presence of potentially biased patterns during the data construction. Through these efforts, we aim to develop the model's tool-using and CoS ability that promotes equitable and unbiased outcomes, fostering trust and inclusiveness in the application of AI systems.
\end{itemize}





\bibliography{anthology,custom}
\bibliographystyle{acl_natbib}
\clearpage
\appendix

\section*{Appendix}
\label{sec:appendix}

\section{Prompt Pattern for ChatGPT Toolkit}
\label{Toolkit_Creation}
\begin{figure*}[!t]
\centering
\includegraphics[width=1.0\textwidth]{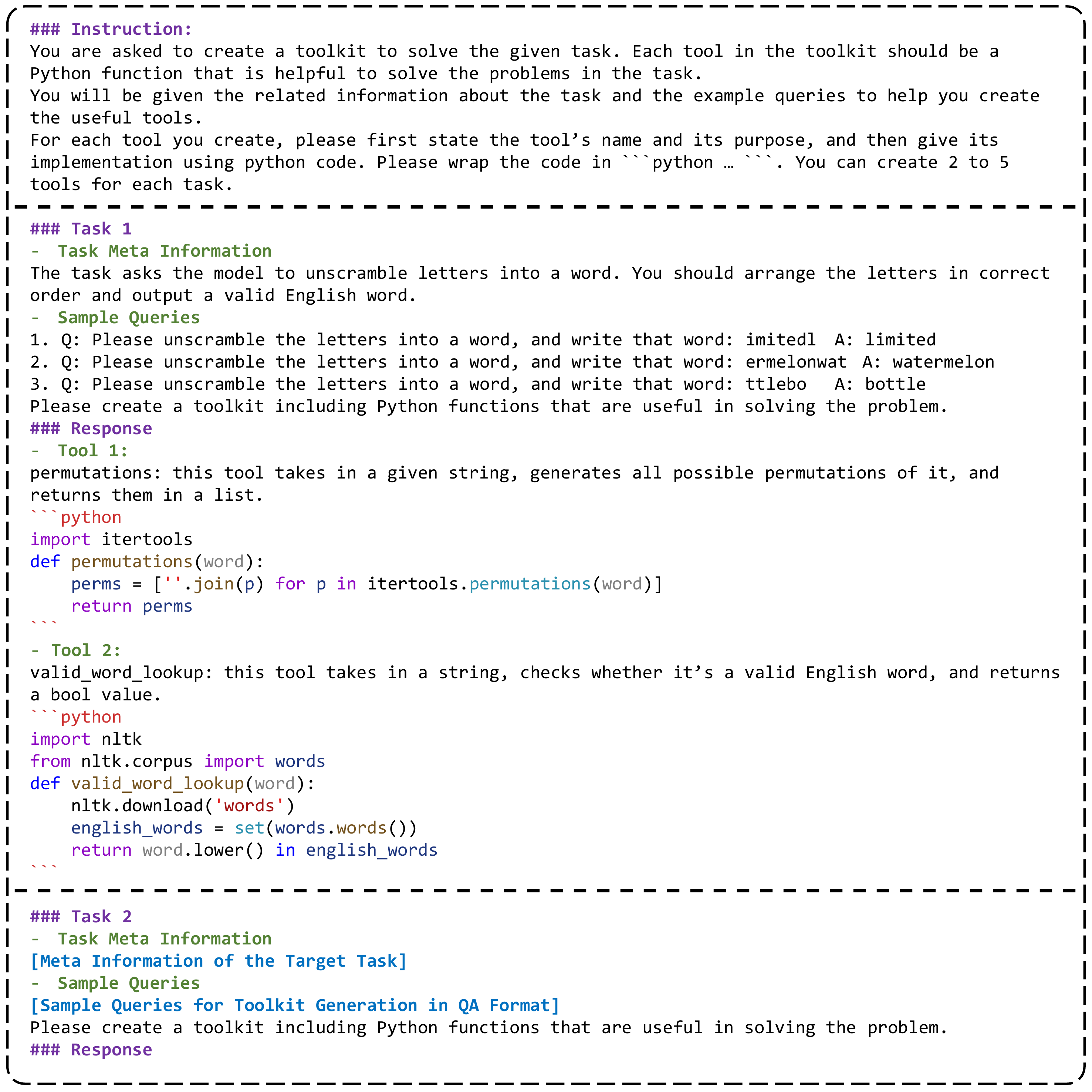}
\caption{The pattern of the prompt given to GPT-3.5-turbo to generate the toolkit.}
\label{fig:apdx_prompt}
\end{figure*}

We show the pattern of the prompt we apply for the creation of toolkits leveraging GPT-3.5-turbo in \Cref{fig:apdx_prompt}. The temperature is set to 0.3 to ensure the model clearly follows the instructions while retaining its creativity to a certain extent. The max length during generation is set to 1024. The prompt shown mainly consists of the instruction for toolkit creation, the demonstration of the format, sample public data, and the task description.

\section{Toolkits for tasks from BIG-bench}
\label{Task_Tools}
\begin{figure*}[!t]
\centering
\includegraphics[width=1.0\textwidth]{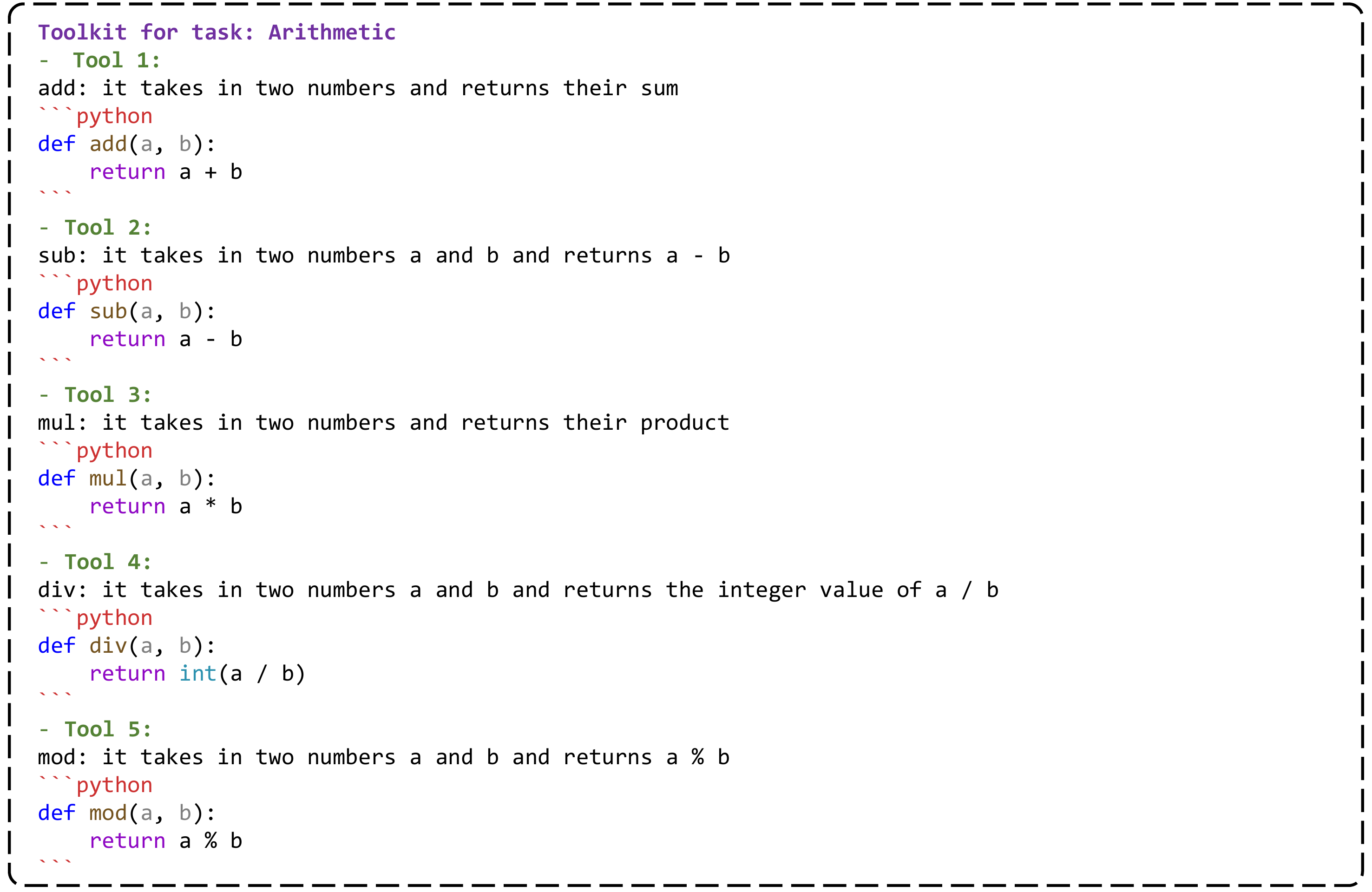}
\caption{The toolkit for task Arithmetic.}
\label{fig:apdx_tool_arithmetic}
\end{figure*}

\begin{figure*}[!t]
\centering
\includegraphics[width=1.0\textwidth]{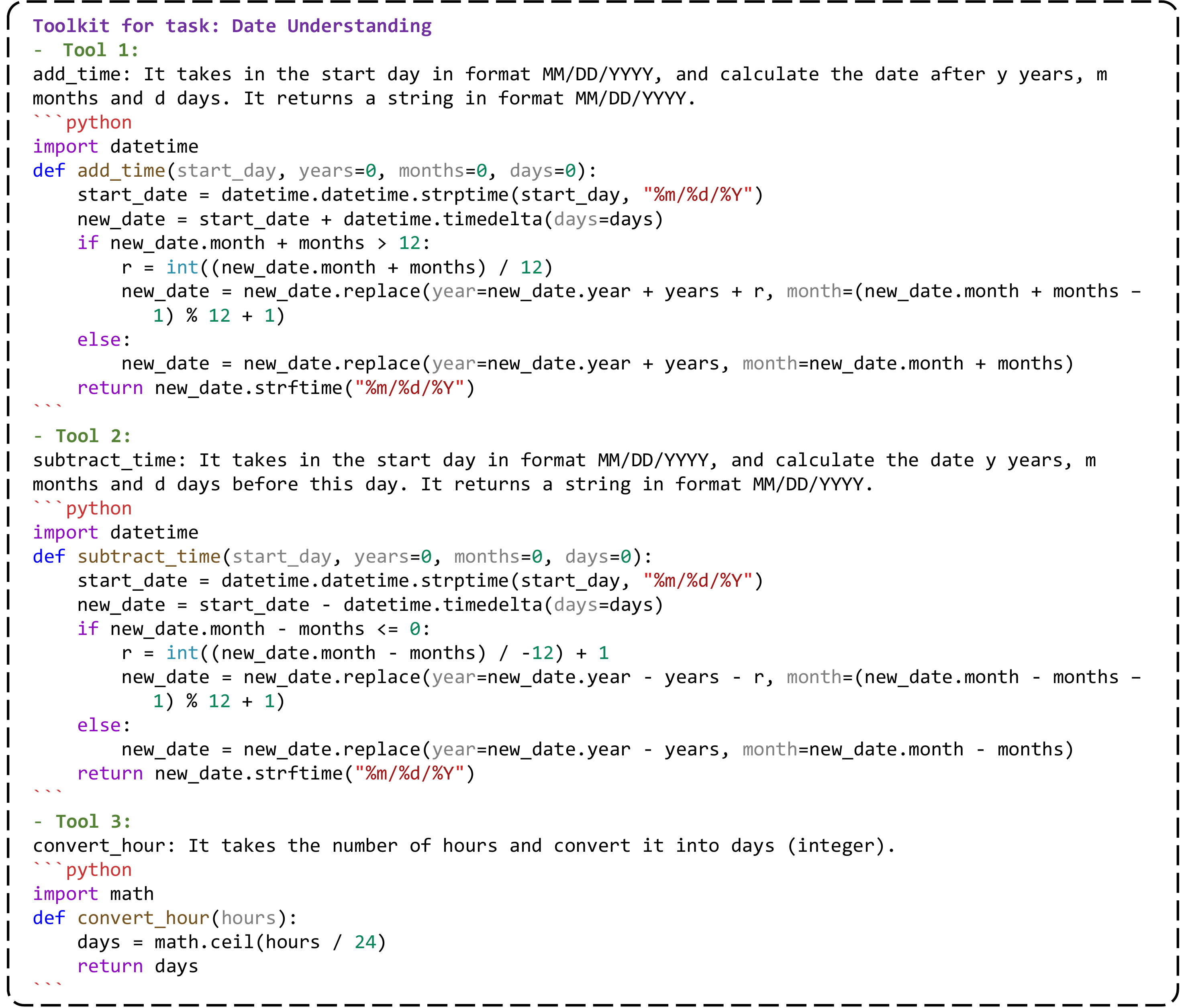}
\caption{The toolkit for task Date Understanding.}
\label{fig:apdx_tool_date}
\end{figure*}

\begin{figure*}[!t]
\centering
\includegraphics[width=1.0\textwidth]{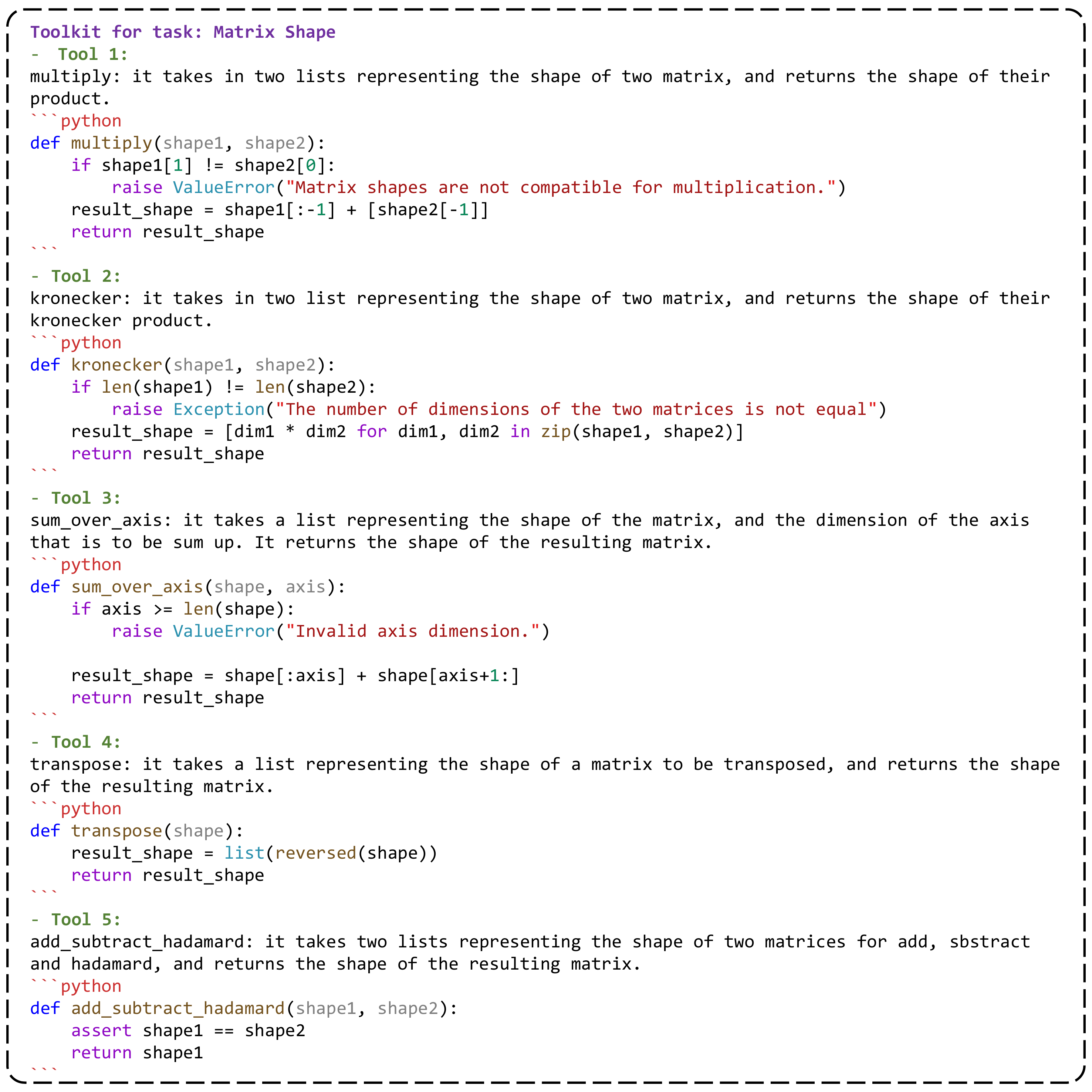}
\caption{The toolkit for task Matrix Shape.}
\label{fig:apdx_tool_shape}
\end{figure*}

\begin{figure*}[!t]
\centering
\includegraphics[width=1.0\textwidth]{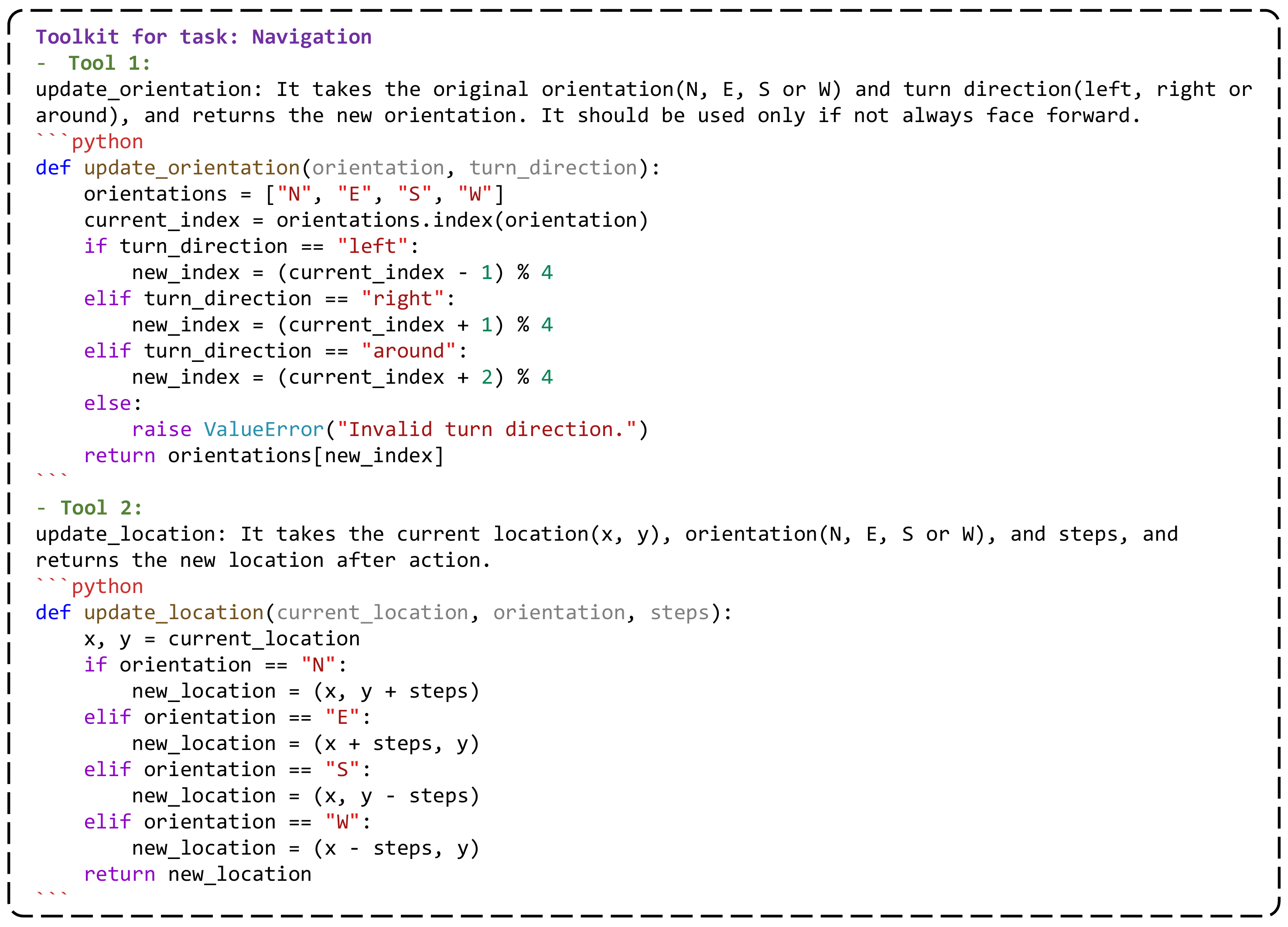}
\caption{The toolkit for task Navigation.}
\label{fig:apdx_tool_navigation}
\end{figure*}

\begin{figure*}[!t]
\centering
\includegraphics[width=1.0\textwidth]{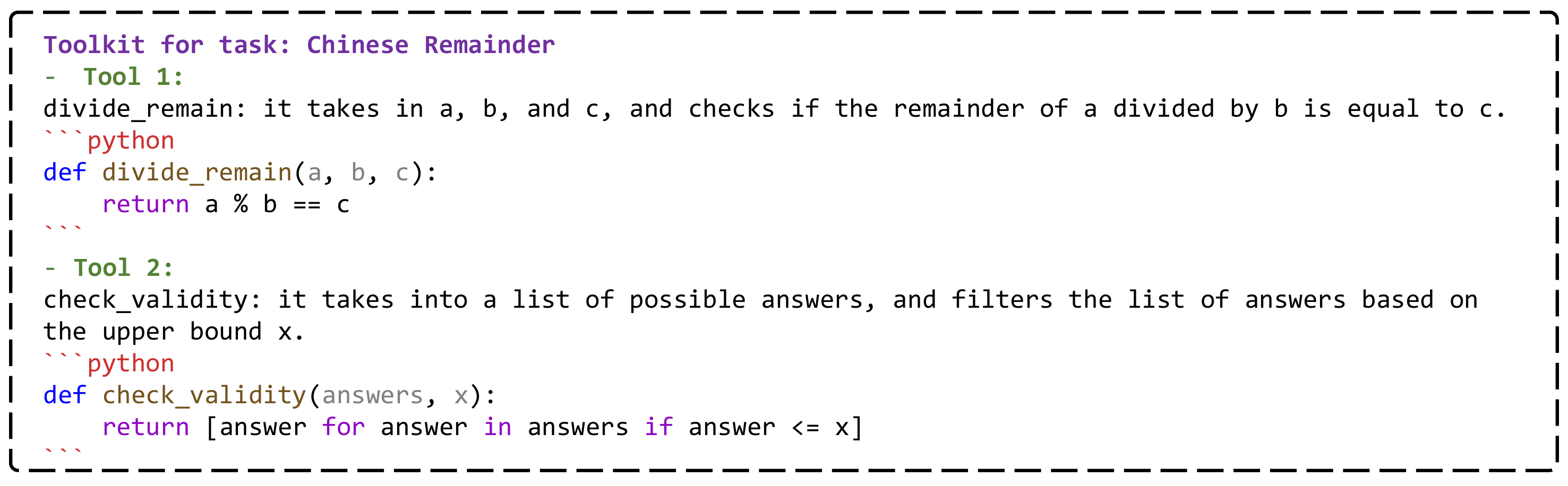}
\caption{The toolkit for task Chinese Remainder.}
\label{fig:apdx_tool_remainder}
\end{figure*}

\begin{figure*}[!t]
\centering
\includegraphics[width=1.0\textwidth]{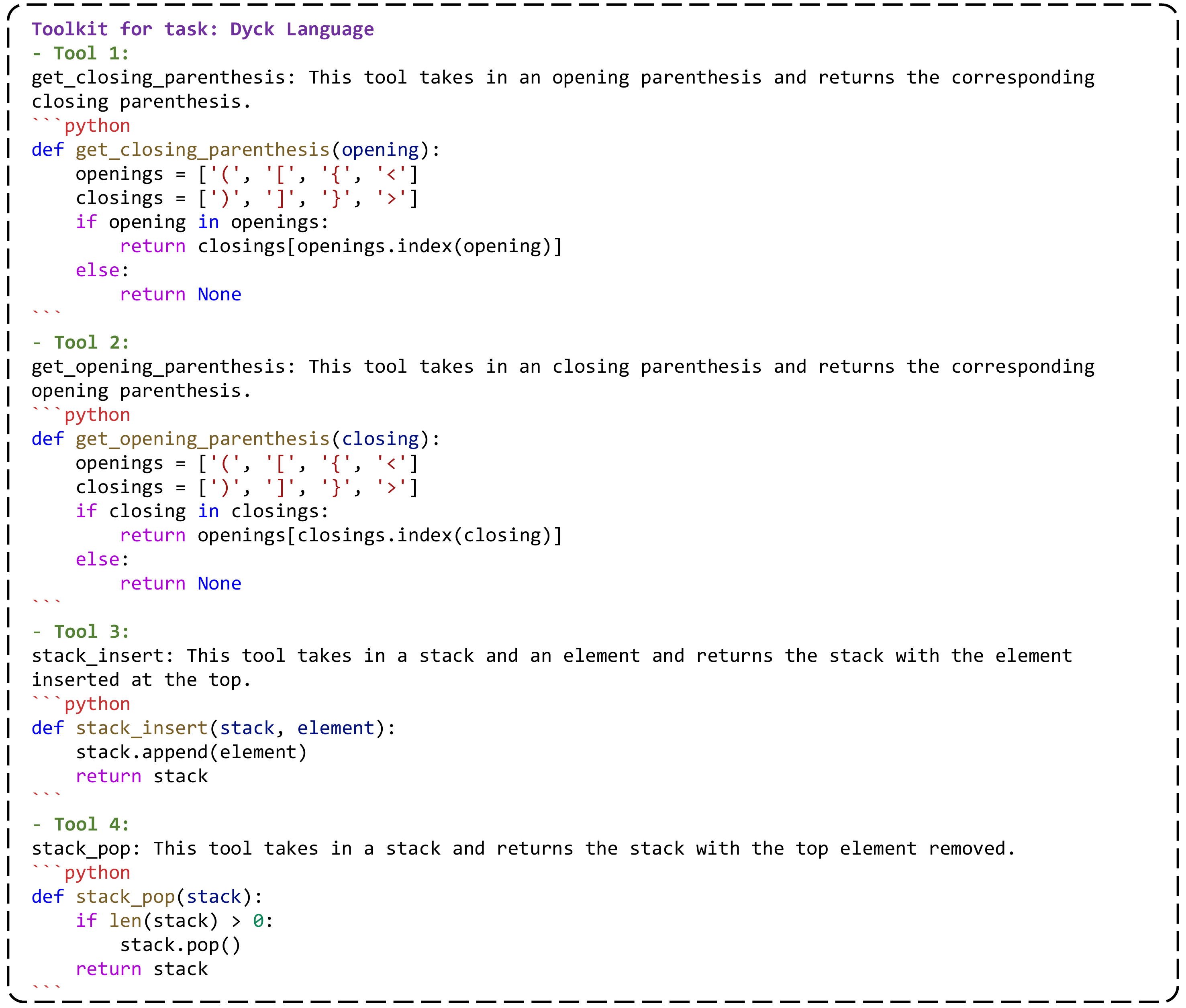}
\caption{The toolkit for task Dyck Language.}
\label{fig:apdx_tool_dyck}
\end{figure*}

\begin{figure*}[!t]
\centering
\includegraphics[width=1.0\textwidth]{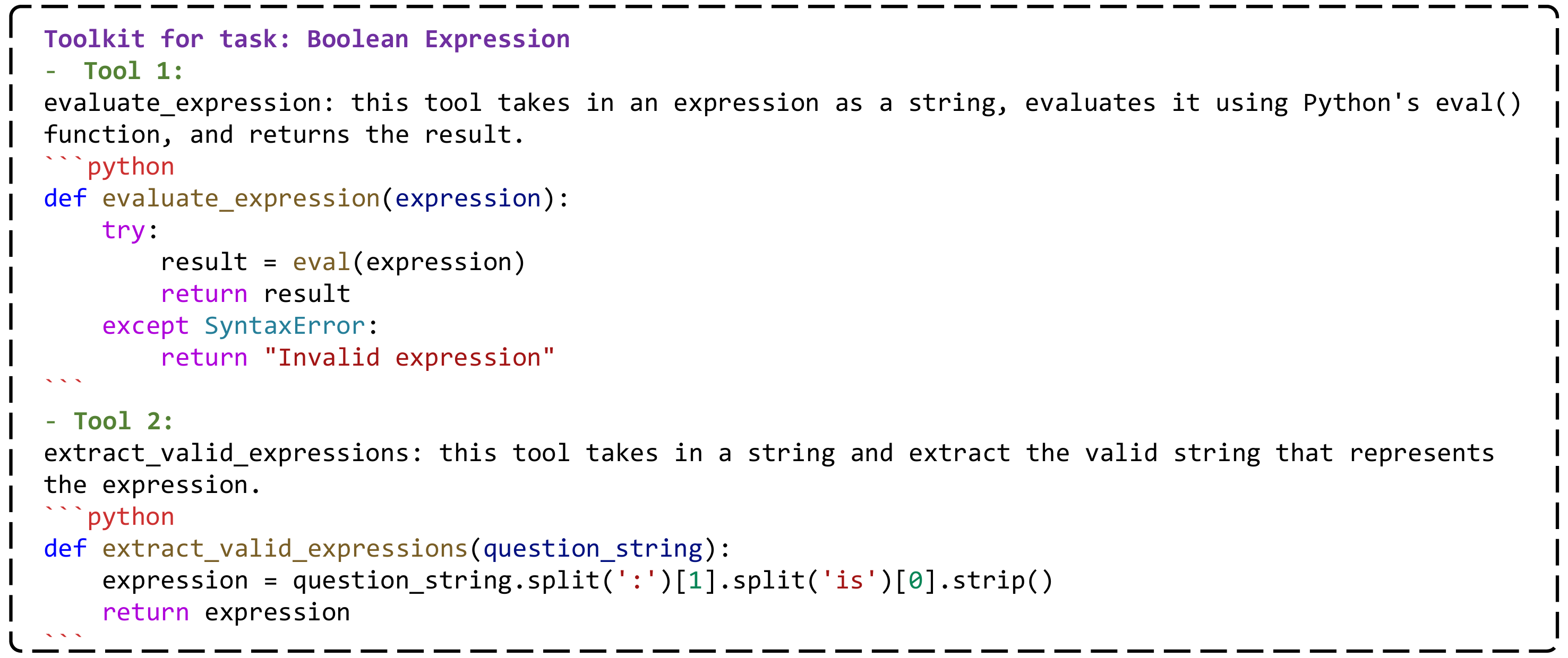}
\caption{The toolkit for task Boolean Expression.}
\label{fig:apdx_tool_boolean}
\end{figure*}

\begin{figure*}[!t]
\centering
\includegraphics[width=1.0\textwidth]{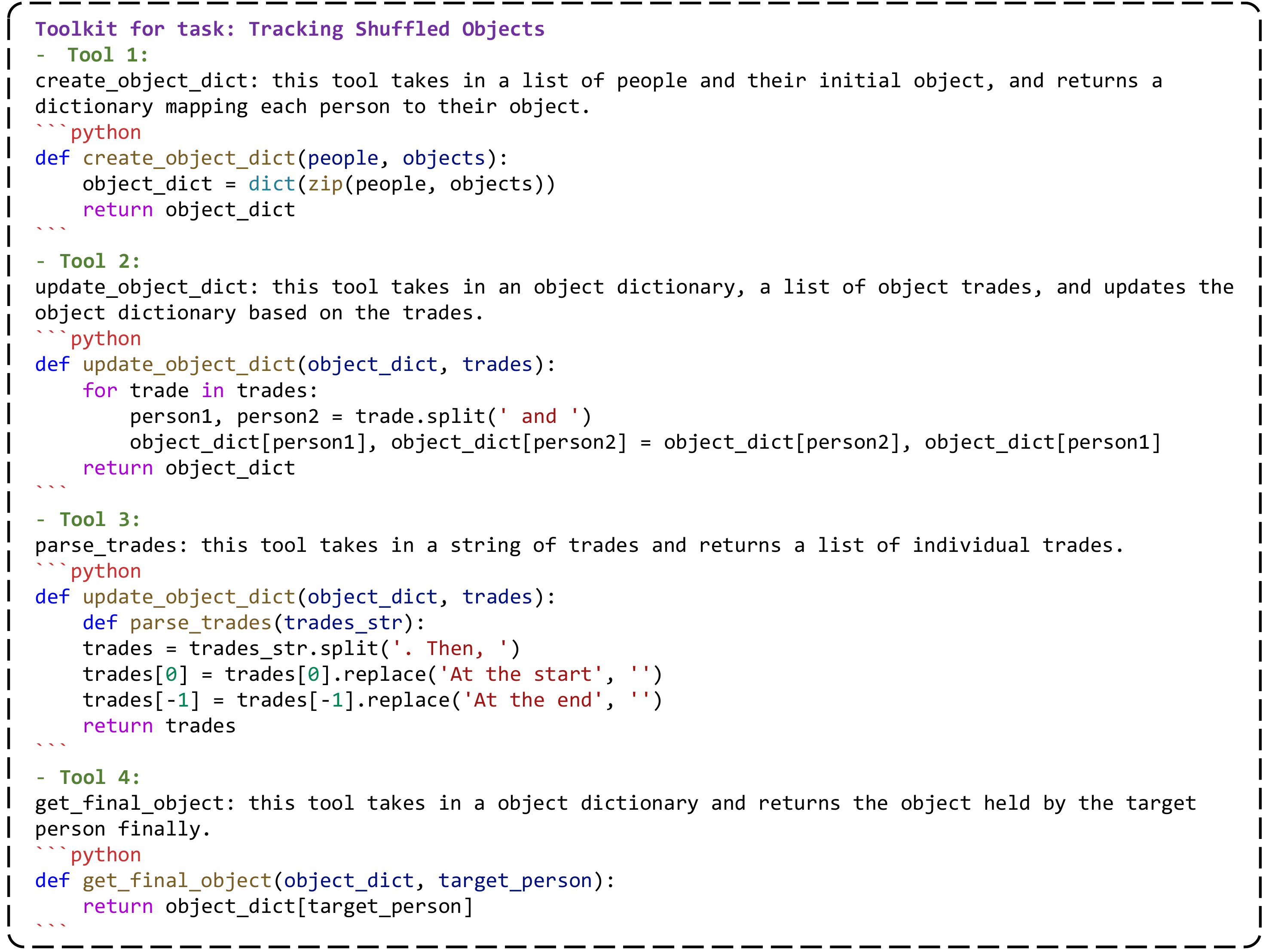}
\caption{The toolkit for task Tracking Shuffled Objects.}
\label{fig:apdx_tool_shuffle}
\end{figure*}

We show in \Cref{fig:apdx_tool_arithmetic,fig:apdx_tool_date,fig:apdx_tool_navigation,fig:apdx_tool_remainder,fig:apdx_tool_shape,fig:apdx_tool_boolean,fig:apdx_tool_dyck,fig:apdx_tool_shuffle} the toolkits that GPT-3.5-turbo created leveraging the prompt mentioned in the previous section. Notice that we show the final version of the toolkit, which may contain certain modifications based on human feedback. For instance, in \Cref{fig:apdx_tool_shape}, we have integrated addition, subtraction, and hadamard operation into one single tool, as all of them do not change the shape of the given matrix. This will effectively reduce the redundant tools and help the model learn with ease.

\section{Settings for Chain-of-Solving}
\label{Composition_Stage_Settings}
\subsection{Choice of Instruction}
\label{Composition_Instruction}
To inspire the models' ability to plan and call the tools during chain-of-solving (CoS), we apply clear instructions to prompt the model. For CoS-Planning, we choose the instruction "\textit{You are presented with a question and several tools that may be useful. Select the useful tools and plan how to solve the problem.}", while for CoS-Calling, we choose the instruction "\textit{Use the tool given in the input to write code to solve the problem.}". This applies to all the settings, including the LLaMA-CoS because it is also tuned in an instruction-following way.

\subsection{Details about Demonstrations}
\label{Composition_Demonstration}
For all the experiments leveraging ChatGPT, despite the instructions, we also provide the model with demonstration examples to showcase the format of planning and calling, as well as to better leverage its potential. The temperature is set to 0.3 during generation, and the max output length is set to 1024.

For the raw LLaMA-7B and Alpaca baselines without being tuned, the demonstration examples are also applied to provide guidance, while the LLaMA-CoS tuned under our {\framework} framework does not need demonstration examples as it is already tuned under the instruction-following paradigm.

\section{Separated Test of CoS-Planning and CoS-Calling}
\label{Separation_Plan_Call}
In {\framework}, planning and calling are combined as a whole CoS process, where the plans generated by the model are again fed back to itself to help guide the generation of the final tool calling decision. To disentangle their functions and better understand their role, we employ tests to measure their accuracy separately.

\begin{figure*}[!t]
\centering
\includegraphics[width=1.0\textwidth]{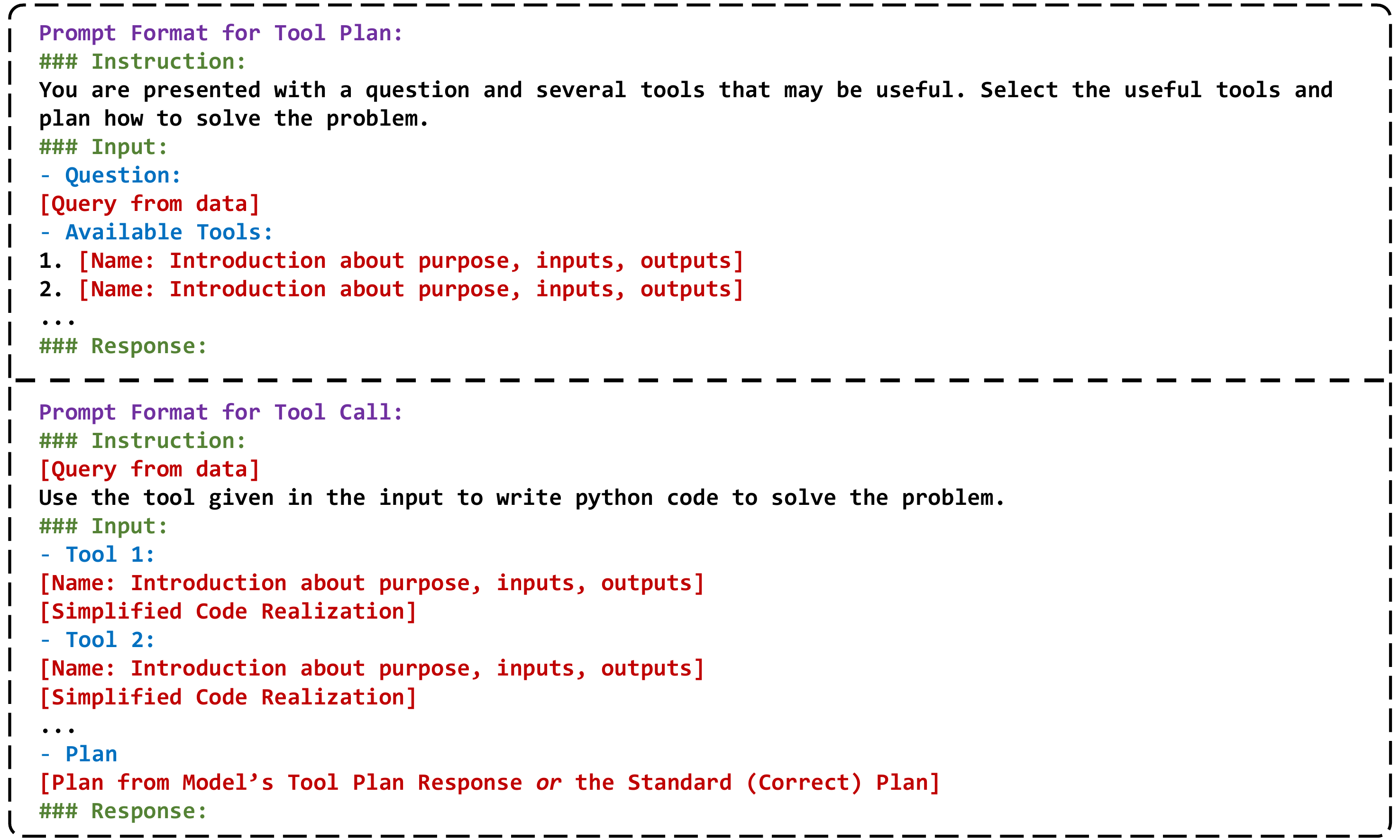}
\caption{The format of the data (and prompt) for CoS-Planning and CoS-Calling.}
\label{fig:apdx_plancall_format}
\end{figure*}

\subsection{CoS-Planning Details}
\label{Tool_Plan_Details}
For the CoS-Planning test, we provide the model with the instructions and all the available tools in the toolkit. In \Cref{fig:apdx_plancall_format}, we showcase the format of the CoS-Planning prompt given to the model.

However, plans are generated in the form of natural language, whose accuracy is hard to measure. For simplicity, we instead only measure if the correct tools are called upon to solve the given problem.

Suppose $K_{T} = \{k_1, k_2, ..., k_N\}$ is the toolkit with $N$ tools for task $T$. For a specific query, we denote the set of useful tools as $K_\texttt{use} \subseteq K_T$ and other redundant tools as $K_\texttt{rdt} \subseteq K_T$. Suppose the set of tools called upon during planning is $K_\texttt{call} \subseteq K_T$, then the correct tools called is denoted as $K_\texttt{correct} = K_\texttt{call} \cap K_\texttt{use}$, and the erroneous tools called $K_\texttt{err} = K_\texttt{call} \cap K_\texttt{rdt}$. These are the exact definitions of the variables that we apply in \Cref{eq_acc_toolplan}.

If all the useful tools are called correctly and precisely, where $K_\texttt{call} = T_\texttt{use}$, the accuracy will be 1.00. Note that this metric is relatively strict because wrong calls will result in a reduction of accuracy.

\subsection{CoS-Calling Details}
\label{Tool_Call_Detailes}
For the CoS-Calling test, the standard (correct) plans will be provided to the model, instead of the plans that the model previously generated. The CoS-Calling test solely aims to investigate the model's ability to follow plans and generate the correct calling decisions. Besides the plans and instructions, only the useful tools with respect to the given query are provided in the prompt, instead of all the tools from the toolkit. We showcase the format of the prompt given to the model in \Cref{fig:apdx_plancall_format}.

The accuracy of CoS-Calling is based on the matching of the model's output to the correct answer. For tasks Arithmetic and Chinese Remainder, the accuracy is evaluated in numerical format; for Matrix Shape, the accuracy is evaluated based on the matching of dimensions list; for all other tasks from BIG-bench, the accuracy is based on the matching of strings between the model's output and the correct answer.

\section{Dataset Construction}
In this section, we provide more details about how CoS-GPT is constructed. We introduce respectively the construction of tool-using data (including planning and calling) and code generation data. All the data points aim to enhance the open-source model's CoS ability.

\subsection{Construction of Tool-Using Data}
\label{Tool-Using_Data_Details}
For each query in AQUA-RAT, GSM8K, and TabMWP, we first utilize ChatGPT to create a diverse set of tools that are potentially relevant to the given query, forming the toolkit. We then provide this toolkit to ChatGPT and allow it to select the most suitable tools. Subsequently, we prompt ChatGPT to generate decision calls based on the selected tools and manually verify the correctness of the resulting outputs. If the final answer is correct, we divide ChatGPT's responses into two distinct components, representing the planning stage and the calling stage, which are then individually added to the dataset. In this manner, the validity of our data points can thus be guaranteed.

Throughout these steps of data construction, we also incorporate demonstration examples sampled from the constructed dataset, thereby expanding the dataset in a self-iterative manner. \Cref{fig:apdx_plancall_format} shows detailed information about the format of the query. Besides the query, we also provide the corresponding CoS-Planning or CoS-Calling response and the implementation of the toolkit with useful and redundant tools.

\subsection{Construction of Code Generation Data}
\label{Code_Generation_Details}
The code generation data in CoS-GPT are sourced from 6 different venues, including Python-Simple, Python-Specific, Math, Algorithm, LeetCode, and Rectification. The objective behind these categories is to enhance the model's proficiency in problem-solving through code utilization, calling existing packages, applying reasoning, employing algorithms, completing codes of challenging competitions, and engaging in self-rectification.

For Python-Simple and Python-Specific, the former aims to boost the models' ability to solve simple problems using codes, while the latter aims to enhance the model's ability to leverage code packages to solve more complex problems. Both these two sets are generated using ChatGPT. We prompt the model with instructions and demonstrations and gather the code snippets the model generated to solve the given problem.

The queries for the Math set are sampled from the training set of MathQA~\citep{amini2019mathqa} and augmented with a code solution based on the given query and reasoning, leveraging ChatGPT. The generated programs are verified to ensure the output answer is the same as the correct one originally, thus ensuring the validity of the augmented data points. The Algorithm set is extracted from the open-source Python algorithm repository, with over 40 categories and more than a hundred diverse algorithms. For each algorithm, we ask ChatGPT to generate a query related to it and use a code snippet to solve the problem. The codes and corresponding queries are then gathered and formed into the instruction-following format.

For the LeetCode set, we directly extract the official open-sourced problems and the code answers from the website and form our data. The Rectification set is gathered from the error codes generated in the five sets before. The error tracebacks and the bad code snippet are fed into ChatGPT, and we leverage it to rectify the codes and generate a correct code snippet that can solve the given query successfully. We gather the generated codes and execute them again, retaining only the ones that give a correct answer finally and form the set based on these valid data points.

\section{Main Experiment Setting Details}
\label{Exp_Setting_Details}
For our main experiment, we finetuned the LLaMA-7B model on four A100-80G GPUs, with a total batch size of 32 and a learning rate of 1e-5. For the model whose performance we demonstrate in \Cref{tab:main_result} and \Cref{tab:main_plancall_results}, its training dataset consists of 1.6K target task-specific data points (8 tasks, 100 for planning and 100 for calling each), 4K tool-using data and 3K code-generation data randomly sampled respectively from the CoS-GPT dataset. We trained the LLaMA-7B on these data for 3 epochs and obtained LLaMA-CoS. 

In addition, for the ablation study about the training on codes we perform in \Cref{Results_Analysis}, we apply 7K tool-using data and remove all the code-generation data points. We keep all the other settings the same in this study.

\section{Case Studies of Diverse CoS Patterns}
\label{diverse_cos_pattern}
In \Cref{fig:apdx_case}, we present three case studies highlighting the diverse nature of LLaMA-CoS in applying planning and calling for tool-using. 

Firstly, LLaMA-CoS exhibits the ability to generate sequential plans involving different tools. In the first case, the model simulates the operation on matrices step by step in a linear way and finally gets the correct result.

Secondly, LLaMA-CoS demonstrates proficiency in executing complex tool calls within branch-loop structures. In the second case, the model learns to use different stack operations based on the character met in the expression, and can call the useful tool in a loop structure.

Lastly, the model showcases its competence in performing nested tool invocations. In the third case, the model is able to directly pass the converted hour retrieved from the previous tool as the input parameter for the next tool, which illustrates a successful nested tool call.

These examples serve to show the robustness, versatility, and adaptability of LLaMA-CoS across a wide range of scenarios.

\section{CoS on Generic Toolkit Details}
\label{Toolkit_Generalization_Details}
We source two new tasks, Dynamic Counting and Unit Interpretation, from the BIG-bench. We apply all the problems in Dynamic Counting for our test of toolkit generalization. However, for Unit Interpretation, we specifically select the data from LV 1 in order for the tools from task Arithmetic to be properly applied. To ensure fairness, we expand the dataset by interactively sampling new questions with similar patterns from ChatGPT and incorporating them until the dataset reaches its original full size. Note that we only aim to showcase the toolkit's generalization ability and compare the performance of LLaMA-CoS and ChatGPT within this work, so we deem expanding the dataset as fair and reasonable under our settings.

\begin{figure*}[!t]
\centering
\includegraphics[width=1.0\textwidth]{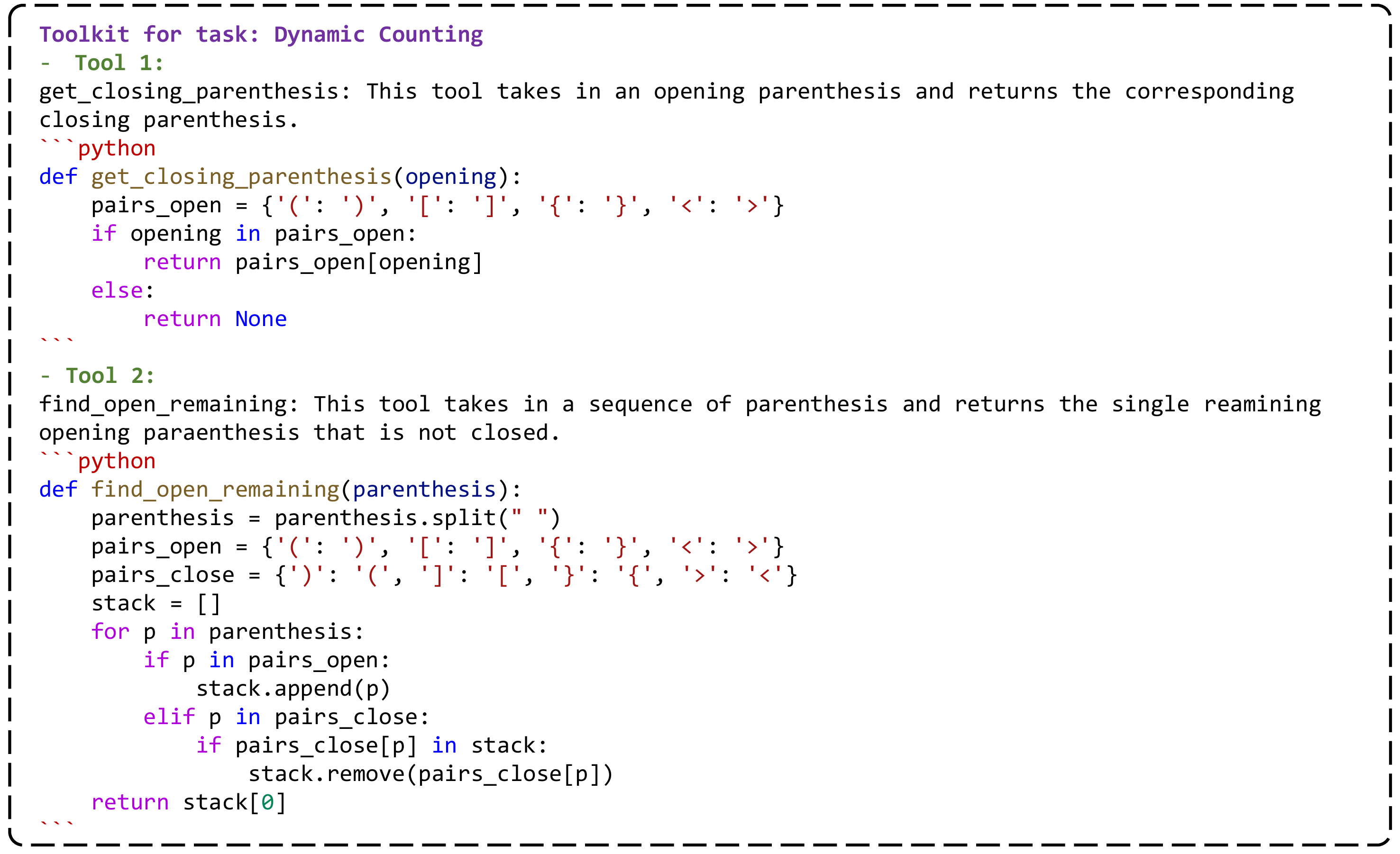}
\caption{The toolkit for task Dynamic Counting.}
\label{fig:apdx_tool_dynamic}
\end{figure*}

\begin{figure*}[!t]
\centering
\includegraphics[width=1.0\textwidth]{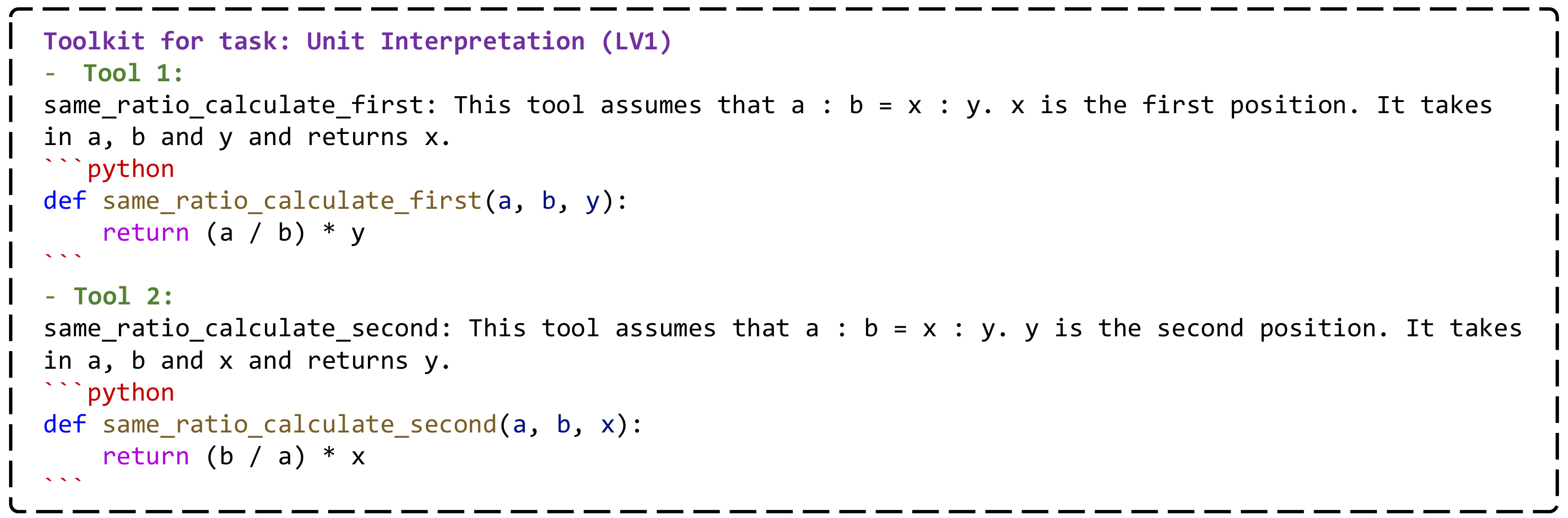}
\caption{The toolkit for task Unit Interpretation (LV 1).}
\label{fig:apdx_tool_unit}
\end{figure*}

We show the toolkits specially tailored for these two new tasks in \Cref{fig:apdx_tool_dynamic,fig:apdx_tool_unit}. The LLaMA-CoS model we apply is still the model we have trained in the main experiment, detailed in \Cref{Exp_Setting_Details}. All the other settings, including the ChatGPT applied under our framework, are kept the same as that in the main experiment.


\end{document}